%% file: main.tex
\documentclass{article} 
\usepackage{iclr2025_conference,times}
\iclrfinalcopy
\input{math_commands.tex}

\usepackage{hyperref}
\usepackage{url}

\usepackage{amsmath}
\usepackage{algorithm}
\usepackage{algpseudocode}
\usepackage{amsfonts}

\usepackage{amsmath}
\usepackage{amssymb}
\usepackage[utf8]{inputenc} 
\usepackage[T1]{fontenc}    
\usepackage{hyperref}       
\usepackage{url}            
\usepackage{booktabs}       
\usepackage{amsfonts}       
\usepackage{nicefrac}       
\usepackage{microtype}      
\usepackage{xcolor}         
\usepackage{multirow}
\usepackage{subfig}
\usepackage{graphicx}
\usepackage{mathtools}
\usepackage{amsthm}
\usepackage{comment}
\usepackage{todonotes}
\usepackage{wrapfig}

\def\N{\mathcal{N}}

\def\A{\mathbf{A}}
\def\E{\mathbb{E}}
\def\R{\mathbb{R}}
\def\B{\mathbf{B}}

\def\x{\mathbf{x}}

\def\0{\mathbf{0}}
\def\1{\mathbf{1}}

\def\I{\mathbf{I}}
\def\L{\mathcal{L}} 
\def\z{\mathbf{z}}

\def\c{\mathbf{c}}

\def\lg{\mathfrak{g}}
\def\lA{\mathbb{A}}

\usepackage[capitalize,noabbrev]{cleveref}
\usepackage{adjustbox}
\usepackage{lineno}

\title{Decafs: Disentangled Conditional adversarial Flows}

\author{
Anirudh Jain and Orion Pharma\\
Aalto University, Finland \\ 
\texttt{anirudh.jain@orionpharma.com}
\And
Sakshi Varshney \\
Aalto University and ARF, India \\
\texttt{sakshi.varshney@airawat.org}
\And
Samuel Kaski \\
Aalto University and University of Manchester, UK \\
\And
Vikas Garg \\
Aalto University and Yai Yai Ltd
}
\begin{document}

\maketitle

\begin{abstract}
Flow-based models have established state-of-the-art performance in generative modeling across domains, but are hard to interpret due to their complex latent embeddings. In particular, the entanglement of generative factors in the latent space hinders controlled generation. We circumvent this issue by appealing to a novel conditional generator based on Lie groups that disentangles an alternative latent space, which is aligned closely with the latent flow space using an adversarial loss.  Our approach facilitates interpretable conditional generation while obviating the need to expand the dimensionality of the flow space (owing to its invertibility requirements).  The proposed model demonstrates strong performance across conditional image (including, outperforming StyleGAN on MNIST, dSprites) and molecule (using standard QM9, ZINC and MOSES) generation tasks.
\end{abstract}

\input{introduction}

\input{related_work}

\input{derivation}

\input{Methodology}
\input{experiments}
\input{conclusion}

\newpage
\input{appendix}

\end{document}

%% file: math_commands.tex
\usepackage{amsmath,amsfonts,bm}

\def\eqref#1{equation~\ref{#1}}

\def\1{\bm{1}}

\DeclareMathAlphabet{\mathsfit}{\encodingdefault}{\sfdefault}{m}{sl}
\SetMathAlphabet{\mathsfit}{bold}{\encodingdefault}{\sfdefault}{bx}{n}

\newcommand{\E}{\mathbb{E}}

\newcommand{\R}{\mathbb{R}}



%% file: introduction.tex
\section{Introduction}
\begin{wrapfigure}[27]{R}{0.3\textwidth}
\centering
\includegraphics[width=4.2cm,height =3.5cm]{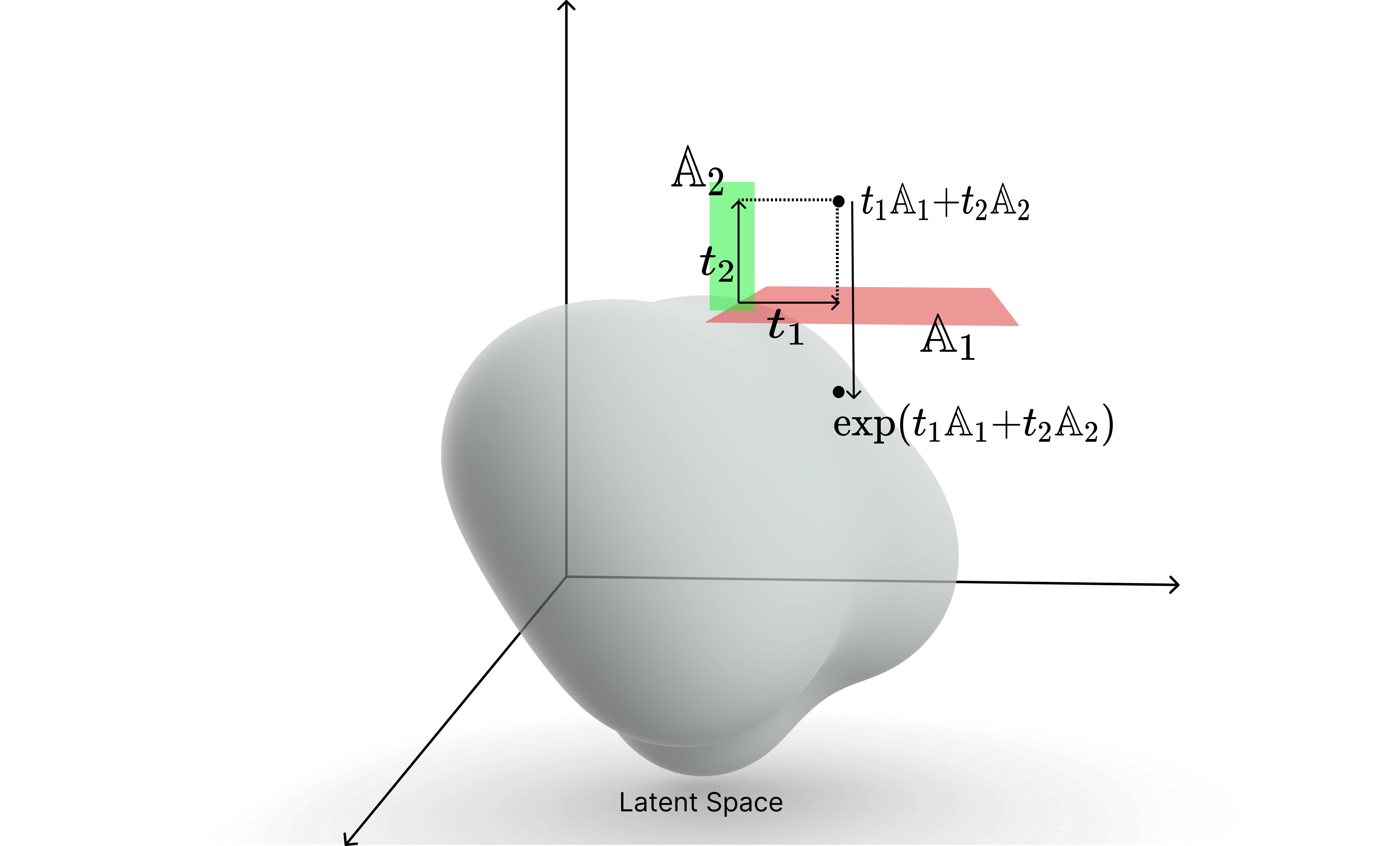}
\caption{\label{fig:lie_generation} Decaf learns disentangled view of flow latent space by modelling each latent vector (equivalently, each point on the manifold) in terms of the coefficients along the Lie group basis $\{\lA_1, \lA_2\}$, representing two independent generative vectors.  Since the resulting vector may be off the manifold,  an exponential map projects this vector back to the nearest point on the manifold. It should be noted that the figure only displays the disentangled subspace of the latent space.} 
\end{wrapfigure}

Recent advances in generative models have enabled some spectacular successes across domains such as text generation, image generation, and drug design \cite{croitoru2023diffusion,verma2023abode,xue2019advances,zhang2023survey}. Methods based on iterative processes such as flows and diffusion models (flow matching), and adversarial methods currently represent the state of the art in generative modelling. Close connections have been established between diffusion models and normalizing flows via probabilistic flow ODEs \cite{song2020score}.

Normalizing flows\cite{dinh2014nice,dinh2016density} are nice because they are invertible and hence the encoding and decoding steps are aligned enabling optimization with exact log likelihood-based objectives (unlike Variational Auto Encoders (VAEs) \cite{kingma2013auto} e.g., that only optimize for a lower bound on the likelihood). Conditional generation with flows however comes at a price, since we need to invert the conditional context together with the unconditional input, expanding the dimensionality of the flow space. Implications include curse of dimensionality, enlarged model size, and further slowing down of the iterative process. Therefore, an overwhelming majority of flow-based methods resort to post-hoc optimizations that typically lead to distribution shifts from the desired conditional distributions.

While Generative Adversarial Networks (GANs) \cite{goodfellow2020generative} have the advantage of not requiring invertible transformations and allowing for more flexible parameterizations, they are not without their drawbacks, including mode collapse and unstable training. 
Consequently, the so-called conditional adversarial flows \cite{liu2019conditional} have been proposed to leverage the best of both worlds, aligning the respective latent spaces. However, the resulting representations learnt using \cite{liu2018constrained} are dense due to the entanglement of the generative factors, which hinders controlled generation. For example, one might want to change only the color of an object, while keeping its shape and size intact.

We advocate an alternative strategy. Rather than growing the architecture of the flows progressively restricted by imposed invertibility constraint, we introduce a novel conditional generator, it is based on Lie Groups, and generates a disentangled latent space aligned with the Flow latent space. 
Since it might not always be possible to completely disengage the different generative factors without losing important information (e.g., in general, we cannot reduce the toxicity of a drug unilaterally without affecting properties such as bioactivity, our generator partitions the latent space into disentangled and coupled parts. Disentangled representations lead to better generalization \cite{lee2021learning, wu2020improving}, efficient sample usage \cite{van2019disentangled}, and explainable representation for generation \cite{chen2016infogan, higgins2017betavae}.
Our approach not only improves interpretability but also enhances control over key properties in tasks ranging from image generation to drug discovery. 
We summarize our contributions as follows:

\begin{itemize}
    \item We propose a novel adversarial conditional flow framework that effectively maps user-specified conditions into a latent space. This latent space is generated by an irreversible generator, which is meticulously aligned with the latent space of a reversible flow. This alignment allows for precise control over the generated outputs based on user-defined conditions, bridging the gap between flexibility and structured generation.

    \item Our approach introduces a Lie algebra-based latent space generator capable of learning both disentangled and coupled components of the aligned flow latent space. This dual capacity enhances the flexibility and interpretability of controlled conditional generation, enabling users to manipulate specific generative factors independently while maintaining the necessary interactions between them. Notably, our work is the first to explore the disentanglement of latent spaces within flow-based models, setting a new direction in the field.

    \item We empirically demonstrate the effectiveness of the proposed \emph{Decaf} model in conditionally generating molecular graphs by manipulating distinct molecular properties, including QED (Quantitative Estimation of Drug-likeness), logP (partition coefficient), and SA (Synthetic Accessibility) score. Our comprehensive evaluations utilize multiple disentanglement metrics in both molecular and image generation domains, providing robust evidence of the model’s superior performance in generating high-quality, conditionally tailored outputs.
    
\end{itemize}

%% file: related_work.tex
\section{Related works}
\paragraph{Conditional generative models}

Normalizing Flows have demonstrated outstanding performance due to their invertible transformations and exact likelihood estimation. However, the invertibility constraint poses a challenge in conditional generation with flows.
 Thus, most of the approaches for conditional generation using Flow-based Generative models \cite{jin2018junction,verma2022modular,shi2020graphaf,luo2021graphdf} use post-hoc property constrained optimization. 
GANs have shown success in conditional generation as they are not restricted by the invertibility constraints and allow floxible transformations to encode the user specified conditions. Conditional GAN \cite{mirza2014conditional}, Auxiliary conditional GAN \cite{odena2017conditional}, InfoGAN \cite{chen2016infogan}, StyleGAN2\cite{karras2020analyzing} are
a few prominent works in the direction of conditional generation using GANs. However, while GANs have good generating quality, mode collapse and unstable training are the challenges faced by GAN based generative models.
Few works \cite{bao2017cvae,xian2019f} combine the strengths of VAEs and GANs to perform conditional generation in applications like zero-shot learning \cite{lampert2013attribute}. 
The recently proposed Conditional Adversarial flow(CAFlow) \cite{liu2019conditional} aims to jointly learn the labels of conditioned images using flow. This model trains an encoder to map conditions to a flow latent space.  However, this work is only limited to the image-based domain and does not explore the disentanglement of the latent space. 
We aim to leverage the strengths of both approaches, combining the flexibility of conditional GANs with the invertibility constraints of normalizing flows, to achieve the best possible results. 
\cite{dhariwal2021diffusion}, \cite{ho2022classifier} proposed a classifier guided and classifier free diffusion model respectively to improve the sample quality of the generated samples using diffusion based models. \cite{sinha2021d2c} proposed a conditional diffusion model for few-shot conditional generation. These approaches does not explore the disentanglement of the latent space for controlled conditional generation.

\paragraph{Disentangled generative models}
A representation is usually considered to be disentangled when a change in one dimension corresponds to a change in one factor of variation \cite{bengio2013representation}. 
Disentangling the generative factors within data is vital for improving controlled conditional generation. It leads to comprehensible, more interpretable, and more manipulable latent representations, which enhance the capability of generative models to produce high-quality outputs conditioned on specific attributes. 

In the domain of image generation, several methods have been proposed to learn disentangled representations with generative models. The most common approaches are based on VAEs with regularizers over the latent space to encourage disentanglement \cite{alemi2016deep, burgess2018understanding, chen2018isolating, esmaeili2019structured, higgins2017betavae,  kim2018disentangling, kumar2018variational, mathieu2019disentangling, rolinek2019variational, rubenstein2018learning}. 
Similar approaches have also been proposed for strings representation of molecules to couple molecular properties with factorized latent dimensions \cite{kang2018conditional, mollaysa2020conditional}. A more recent formulation for VAEs is using Lie groups to parameterize the encoder for disentangled representations for images \cite{zhu2021commutative} and molecular graphs \cite{mercatali2022symmetry}. 
One class of models is based on the GAN framework that regularizes a subset of latent factors of the generator to be correlated with variations in data \cite{chen2016infogan, jeon2021ib,  lin2020infogan, nie2020semi}. 
\cite{yang2023disdiff,wu2023uncovering} 
Due to the invertibility constraints inherent in flow-based models, there has been very little research focused on disentangling the flow latent space. To the best of our knowledge, our work is the first to explore disentangling the latent space in flow-based models. We propose a irreversible novel conditional generator capable of generating disentangled latent space which is closely aligned to the reversible Flow latent space.

%% file: derivation.tex
\section{Proposed Framework}

\subsection{Controlling Generation}
Below we use the shorthand $p_{\theta}(x) = p(x;\theta)$. Let $p_{data}$ denote the true data distribution. 

Consider the flow objective.

\begin{eqnarray*}
\min_{p_\theta} & KL(p_{data}(x)||p_{\theta}(x))\\
=  \max_{p_\theta} & \mathbb{E}_{\x \sim p_{data}} \log p_\theta(\x) \\   
=  \max_{p_\theta} & \mathbb{E}_{\x \sim p_{data}} \mathbb{E}_{\z \sim p_\theta} \, \left( \log p_\theta(\z) + \log \left| \frac{\partial \z}{\partial \x} \right| \right) 
\end{eqnarray*}

Here, $z \sim p_\theta$ is obtained by applying a sequence of invertible transformations of the form 
$$\z = f_\theta(\x) = f_1 \circ \cdots \circ f_T(\x)~.$$

$f_{\theta}$ is usually a complex mapping, resulting in complex, highly entangled latent embeddings, hindering controlled generation. Therefore, we seek an alignment between $f_{\theta}$ and a simpler model  $g_{\phi}$  such that: (1) $p_{\theta}(z) \approx p_{\phi}(z)$,  (2)  $f_{\theta}$ and $g_{\phi}$ produce latent spaces that yield similar posterior class  distributions; i.e.,  $p_{\theta}(c|z) \approx p_{\phi}(c|z)$, and (3) $g_{\phi}$ yields a latent space that can be disentangled to enable controlled generation. Note that (1) and (2) collectively imply that   
$p_{\theta}(z, c) \approx p_{\phi}(z, \c)$, leading to joint alignment of the latent spaces as well as the conditional contexts; in particular, affording $p_{\theta}(z|\c) \approx p_{\phi}(z|\c)$.  



We have flexibility in terms of how (1) can be (approximately) achieved; e.g., we could use a GAN or VAE for $G_{\phi}$. We opt for Wasserstein GAN here. For (2) we can jointly train an additional parameterizable prediction component $p_{\psi}$ that takes a latent embedding $z$ and predicts the conditioning vector $c$; i.e., we impose a loss to encourage $p_{\psi}(\c|f_{\theta}(x)) \approx p_{\theta}(\c|f_{\theta}(x))$ and  $p_{\psi}(\c|g_{\phi}(x)) \approx p_{\phi}(\c|g_{\phi}(x))$. For (3) we appeal to Lie groups. 

Note that our approach can be readily leveraged to enable controlled generation for pretrained flow models as well. Moreover, while we focus here on flow models, in principle, our approach can be used to perform controlled generation with related iterative generative models such as diffusion models and flow matching models.

\subsection{Disentangling Latent space}

Here we aim to define disentanglement of the generated flow latent space. Disentanglement refers to the ability to separate different underlying factors of variation in data, which can enhance interpretability and improve controlled generation.

%% file: methodology.tex
\section{Method }

\begin{figure*}[t]
\vskip -0.1in
\begin{center}
\centerline{\includegraphics[width=\linewidth]{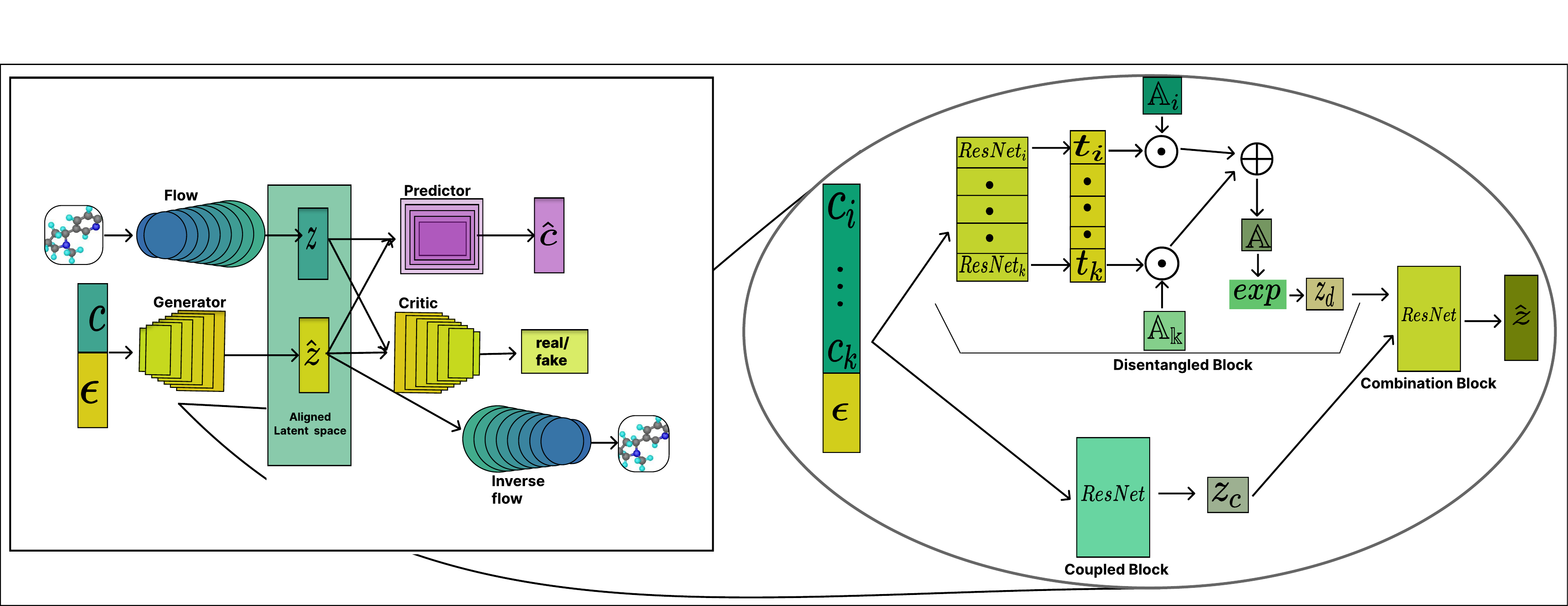}}
\caption{\textbf{Left:}\textbf{Decaf Architecture}, comprises of four primary blocks: A \textbf{Flow network} which learns an invertible transformation between true data distribution and the latent space, A \textbf{Lie-group generator} which generates a latent space conditioned on the input labels, A \textbf{discriminator Network} that discriminates the flow latent space and the generated latent space for adversarial training, A \textbf{ predictor network} that aids the flow and generator network to form clusters in the latent space based on the input labels
\textbf{Right:} The architecture of Lie-Group Generator, comprises of a \textbf{coupled block} which takes as input condition $c$ and noise vector $\epsilon \sim \mathcal{N}(0,I)$ and generates coupled latent vector $z_c$, and a \textbf{disentangled block} which comprises of independent Resnet to predict coordinates $t_i$ for each corresponding input label and combines them with the Lie basis $\lA$ to generate the disentangled latent space $z_d$, a \textbf{combination block} that combines the coupled and disentangled latent spaces to generate the final output $\hat{z}$. Figure is best viewed in color.}
\label{fig:model_architecture}
\end{center}
\vskip -0.3in
\end{figure*}

\subsection{Decaf: Disentangled conditional adversarial flow}
We define the underlying idea and comprehensive design of our proposed method, \textbf{\emph{Decaf}}, as depicted in Figure~\ref{fig:lie_generation}, in this section.
\emph{Decaf} contains a \textbf{multi-scale reversible flow} which generates the data samples, a \textbf{Lie-group generator} which learns to generate conditional latent space aligned with flow space, a \textbf{discriminator} network which discriminates Lie-generator and flow latent spaces, and a prediction network that aids in cluster formation using conditional information. 
\emph{Decaf} first trains a flow in an unconditional fashion with the objective defined in eqn.~\ref{eqn:flow}. 
\paragraph{Flow}$\mathcal{F}$ learns a bijective mapping $z=f_\theta(x)$ from the true data distribution to the latent space using a sequence of invertible transformations. $z$ belongs to a prior distribution $p(z)$, which follows standard Gaussian distribution $\mathcal{N}(0,I)$. Normalizing flow is optimised using the exact log likelihood-based objective function. Hence the loss for training the flow is as follows:
\begin{equation*}
    \mathcal{L}_{flow} = - \mathop{\mathbb{E}}_{z \sim p(z)} \left[\log p(\mathbf{z}_i) + \log \left|\det \left(\frac{\partial \mathbf{f}}{\partial \mathbf{x}}\right)\right|\right]
\label{eqn:flow}
\end{equation*}
\paragraph{Lie group generator}\textbf{G} learns to generate conditional latent space aligned with the flow space. The detailed description of the generator network is demonstrated in Figure.~\ref{fig:model_architecture}
It generates a latent space vector $\hat{z} = G_\theta(c,\epsilon)$ from the condition vector $c$ and the noise vector $\epsilon \sim \mathcal{N}(0, I)$ as input.
\emph{Decaf} enforces disentanglement in the generator for improved conditional generation as entanglement in the 
generative factors can impede the controlled generation. 
However, generative factors can have correlations with each other that will not be captured by the disentangled representation. Therefore, \emph{Decaf's} Generator consist of a block for producing coupled latent factors $z_c$ and a block for producing disentangled latent factors $z_d$. 
Both the entwined and the disentangled components contribute to the final latent vector $\hat{z}$.
 The generator network imposes the disentanglement induced by Lie algebra. The disentangled latent space $z_d$ is generated using a Lie Group $\mathcal{G}$ and a Lie algebra $\lg$. The disentangled block in the generator network first learns the Lie algebra basis element $\{\lA_i\}_{i=1}^k\in \lg $, for each coordinate $t_i$. The  disentangled latent vector $z_d$ is obtained by:
\begin{equation*}
    \z_d = \exp ( \sum_{i=1}^k t_i \lA_i )
\end{equation*}
 where each $t_i=m_i(c_i)$ is obtained via an independent neural network $m_i$ that takes the corresponding condition label as input.
The coupled latent vector $z_c$ is generated using a block which takes condition $\c$ and noise vector $\epsilon$: 
 \begin{equation}
     \z_c = g(c, \epsilon)
 \end{equation}
The final latent vector is generated by combining the coupled and disentangled latent vectors as follows:
 \begin{equation}
     \hat{\z} = h(\z_c, \z_d) 
 \end{equation}
where $g$ and $h$ are parameterized by neural networks. 

\paragraph{Discriminator}network \textbf{$D$} learns to differentiate the flow latent space $\z$ to the generated latent space $\hat{z}$. 
The Lie group generator $G$ and the discriminator network $D$ constitute a \textbf{GAN} framework and are trained in an adversarial manner. The generator is trained to be aligned with flow latent space. We use WGAN objective\cite{arjovsky2017wasserstein} to train the generator $G$ and discriminator $D$, The GAN based objective function is defined as follows:

\begin{multline}
    \L_{\text{GAN}} = \min_G \max_D \E_{\x, c\sim \mathcal{D}_{\text{M}}}\underbrace{[D_{\Omega} (f_\theta(\x))]}_{\text{Flow latent space as real data}}  -  \E_{\x, c\sim \mathcal{D}_{\text{M}}} \E_{\epsilon \sim p(\epsilon)} \underbrace{[D_\Omega (G_\phi(c, \epsilon))]}_{\text{Lie generator output as fake data}}  \\ + \lambda_{GP} \E_{\z \sim \N(\0, I)} \underbrace{\left[( \| \nabla_z D(z) \|_2 - 1 )^2 \right]}_{\text{Gradient Penalty}}  
\end{multline}

\paragraph{Predictor network}$\mathcal{\textbf{P}}$ is a classifier/regressor network that learns to classify the flow latent space $z$ and the generated latent space $\hat{z}$. The predictor network aids the flow network and the generator network to learn the class posterior probabilities. The loss function used for the predictor network is as follows:
\begin{equation}
    \mathcal{L}_{Pred} = \E_{z \sim p(z), \hat{z} \sim G(c,\epsilon)} {\vert\vert P_{\psi}(z)-\c\vert\vert}^2 +  {\vert\vert P_{\psi}(\Hat{z})-\c\vert\vert}^2
\end{equation}
\paragraph{} The final \textbf{objective} used for the training of \emph{Decaf} is as follows:

\begin{equation}
     \mathcal{L}_{Decaf} = \lambda_F\mathcal{L}_{flow} +\lambda_G \mathcal{L}_{GAN} + \lambda_P\mathcal{L}_{Pred} 
     + \lambda_H \L_{Hess} + \lambda_C \L_{Comm}
\end{equation}
    
We empirically select the values of $\lambda$, the corresponding values are provided in the \ref{appendix:img_gen_model}

The comprehensive algorithm~\ref{alg:training} provides the training procedure for \emph{Decaf}, while algorithm~\ref{algo: generation} describes the conditional generation.

\begin{algorithm}
\caption{Training: Decaf}
\begin{algorithmic}[1]

\State \textbf{Input:} Dataset $D = \{x_i, c_i\}_{i=1}^N$, Pre-trained Flow $F$, Noise vector $\epsilon \sim \mathcal{N}(0, I)$, Iterations $n_{\text{iter}}$, Batch size $B$, Batches $n_b$ 
\State \textbf{Initialize:} Predictor parameters $\psi$, Load Pre Trained Flow parameters $\theta$ 

\State \textbf{Step 1: Train Flow and Predictor}
\For{$i = 1$ to $n_{\text{iter}}$}
    \For{$j = 1$ to $n_b$}
        \State Sample $\{x_1, c_1\}, \dots, \{x_B, c_B\}$
        \State Compute latent codes $z_b = F_{\theta}(x_1, \dots, x_B)$
        \State $\mathcal{L}_{\text{flow}} = \frac{1}{B} \sum_{i=1}^B \left[ \log p(z_i) + \log \left| \det \left( \frac{\partial f}{\partial x} \right) \right| \right]$
        \State $\mathcal{L}_{\text{Pred}} = \frac{1}{B} \sum_{i=1}^B \|\mathcal{P}_{\psi}(z) - c \|^2$
    \EndFor
    \State Update $\theta$, $\psi$ using respective optimizers.
\EndFor

\State \textbf{Step 2: GAN Training}
\State \textbf{Initialize:} Generator parameters $\Omega$, Discriminator $\phi$

\For{$i = 1$ to $n_{\text{iter}}$}
    \For{$j = 1$ to $n_b$}
        \State Sample latent noise $z = F(x)$
        \State Generate $\hat{z}$$ = G_{\Omega}(c, \epsilon)$
        \State $\mathcal{L}_{\text{GAN}} = \frac{1}{B} \sum_{i=1}^B \left[ -D_{\phi}(z) + D_{\phi}(\hat{z}) \right]$
        \State $\mathcal{L}_{\text{Comm.}} = \frac{1}{T} \sum_{i,j} (\mathbb{A}_i \mathbb{A}_j - \mathbb{A}_j \mathbb{A}_i)$
        \State $\mathcal{L}_{\text{Hess.}} = \frac{1}{T} \sum_{i,j} (\mathbb{A}_i \mathbb{A}_j)$
    \EndFor
    \State Update $\Omega$, $\phi$, $\psi$ using respective optimizers on combined loss:
    \State $\Omega \leftarrow optimizer (\frac{1}{n_b}\sum_{b=1}^{n_b} (\mathcal{L}_{GAN} + L_{Pred} + \lambda_H \mathcal{L}_{Hess.} + \lambda_C \mathcal{L}_{Comm.}),\Omega)$
    \State $\phi \leftarrow optimizer (\frac{1}{n_b}\sum_{b=1}^{n_b}\mathcal{L}_{GAN},\phi)$
    \State $\psi \leftarrow optimizer (\frac{1}{n_b}\sum_{b=1}^{n_b} \mathcal{L}_{pred},\psi)$
    
\EndFor

\end{algorithmic}
\end{algorithm}

\begin{algorithm}[ht] \label{algo: generation}
\caption{Generation with   \emph{Decaf}}
   \label{alg:generation}
\begin{algorithmic}[1]
   \State {\bfseries Input:} property vector $c$,  Noise vector $\epsilon \sim \mathcal{N}(0,I)$ 
   \State Sample $\hat{z}=G(c,\epsilon)$ 
   \State $\hat{x}=F^{-1}_\Omega(\hat{z})$
   \end{algorithmic}
\end{algorithm}

%% file: experiments.tex
\section{Experiments}
We employ widely used benchmark datasets to empirically evaluate \emph{Decaf} performance on controllable conditional generation provided user-specified inputs and we further evaluate the quality of disentangled generation. We investigate the performance of \emph{Decaf} on images as well as the crucial molecular graph domains. 
 We compare the performance of our model with other prominent works in the direction of conditional and disentangled generation. We show that \emph{Decaf} provides state-of-the-art results for disentangled conditional generation using Normalising flows. We also demonstrate the effectiveness of \emph{Decaf} on conditional generation for 2D Molecular graphs with high validity and novelty, results are provided in Appendix \ref{appendix:molecular_arch}. All models are trained on single Tesla V100 GPU with average wall time of 4 hours.

\subsection{Image generation}
\paragraph{Datasets} 

We evaluate   \emph{Decaf} for conditional image generation on 
MNIST\cite{lecun1998mnist} and dSprites\cite{higgins2017betavae} dataset. 
MNIST dataset is composed of 60,000 training images each of size $28 \times 28$ with one hot class label as explicit conditions.
We perform quantitative analysis of disentanglement on dSprites dataset; which consists of 737,280 binary 2D images of size $64 \times 64$  with shapes (square, heart and oval) along with size (6 values), rotation (40 values over the $2\pi$ range), X position (32 values) and Y position (32 values). All 5 factors are independent and each image has a unique combination of factors. The one-hot encoding over each factor is used as input for controllable generation.

\paragraph{Metrics} 

We use Fréchet inception distance (FID)\cite{heusel2018gans} metrics to quantitatively evaluate the generation quality of the image dataset (MNIST ). FID score measures how similar the generated images are to the real images. The lower the FID attained the better the generation quality of the defined model.

Further, to quantitatively evaluate the disentanglement in the conditional generation of   \emph{Decaf}, we rely on various popular metrics such as FactorVAE Metric(FVM) \cite{kim2018disentangling}, Disentanglement, Completeness and Informativeness (DCI)\cite{eastwood2018a}, Seperated Attribute Predictability (SAP) score \cite{kumar2017variational} and Mutual Information Gap (MIG) \cite{chen2018isolating}. 
Appendix~\ref{appendix:dis_metrics} contains the detailed description of the above mentioned metrics.

\label{para:img_model}
\paragraph{Model}We train a multi-scale flow architecture based on the state-of-the-art Flow++ model \cite{ho2019flow} for both MNIST and dSprites dataset generation.
 We employ a residual architecture-based\cite{he2015deep} predictor network to encourage clustering in the latent space based on the provided conditions. 
 The \emph{Decaf} critic and predictor network is built upon Residual networks.
Our proposed model efficiently generates conditional outputs while carefully managing additional parameters. It requires only 10\% more parameters to achieve disentangled conditional generation. 

Adam optimizer is used to train each model, and the Appendix~\ref{appendix:img_gen_model} contains information on the detailed architecture and hyper-parameters.

\begin{table}[]
\subfloat[] 
{\begin{tabular}{@{}lll@{}} 
\toprule
Model     & FID ($\downarrow$)  & DCI ($\uparrow$)  \\ \midrule
BetaVAE$^*$ & 107.33 & 0.87 \\
FactorVAE$^*$ & 46.87  & 0.85 \\
StyleGAN$^\circ$ & 13.21  & -    \\
DDPM$^\#$ & 3.67 & - \\
Flow (Ours)     & 16.88  & -    \\
\textbf{Decaf (Ours) }  & 10.80  & 1.00 \\ \bottomrule
\end{tabular} \label{tab:mnist}}
\hspace{0.1in}
\subfloat[]
{\begin{tabular}{@{}lllll@{}}
\toprule
Model     & DCI(($\uparrow$) & MIG($\uparrow$)  & SAP($\uparrow$) & FVM($\uparrow$) \\ \midrule
BetaVAE$^*$ & 0.41       & 0.21 & 0.55 & 0.63 \\
FactorVAE$^*$ & 0.74       & 0.43 & 0.56 & 0.83 \\
LieVAE$^*$    & 0.23       & 0.30 & 0.54 & 0.86 \\
InfoGAN$^*$  & 0.71       & 0.37 & 0.58 & 0.88 \\
IB-GAN$^*$  & 0.67       & -    & -    & 0.80 \\
Info-StyleGAN$\circ$ & - & 0.27 & - & 0.84 \\
\textbf{Decaf (Ours)}    & 1.00 & 0.78 & 0.50 & 1.00 \\ \bottomrule
\end{tabular}\label{tab:dsrites}}
\caption{\textbf{(a)}: FID and DCI scores on MNIST image generation from latent vectors. Results marked with ($*$) are obtained from \cite{ngo2022transitive}, $\#$ from \cite{pirhayatifard2023improving} \textbf{(b)}: Disentanglement metrics of \emph{Decaf} and baselines on dSprites dataset. Results marked with ($*$) are obtained from \cite{lin2020infogan} and ($\circ$) from \cite{nie2020semi}}
\vspace{-0.2in}
\end{table}

\begin{figure}[ht]
\includegraphics[width=0.5\linewidth,height=4.6cm]{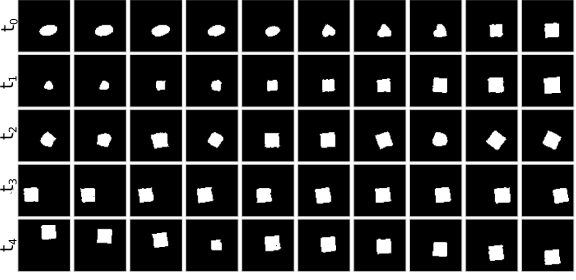}
    \quad
    \includegraphics[width=0.5\linewidth]{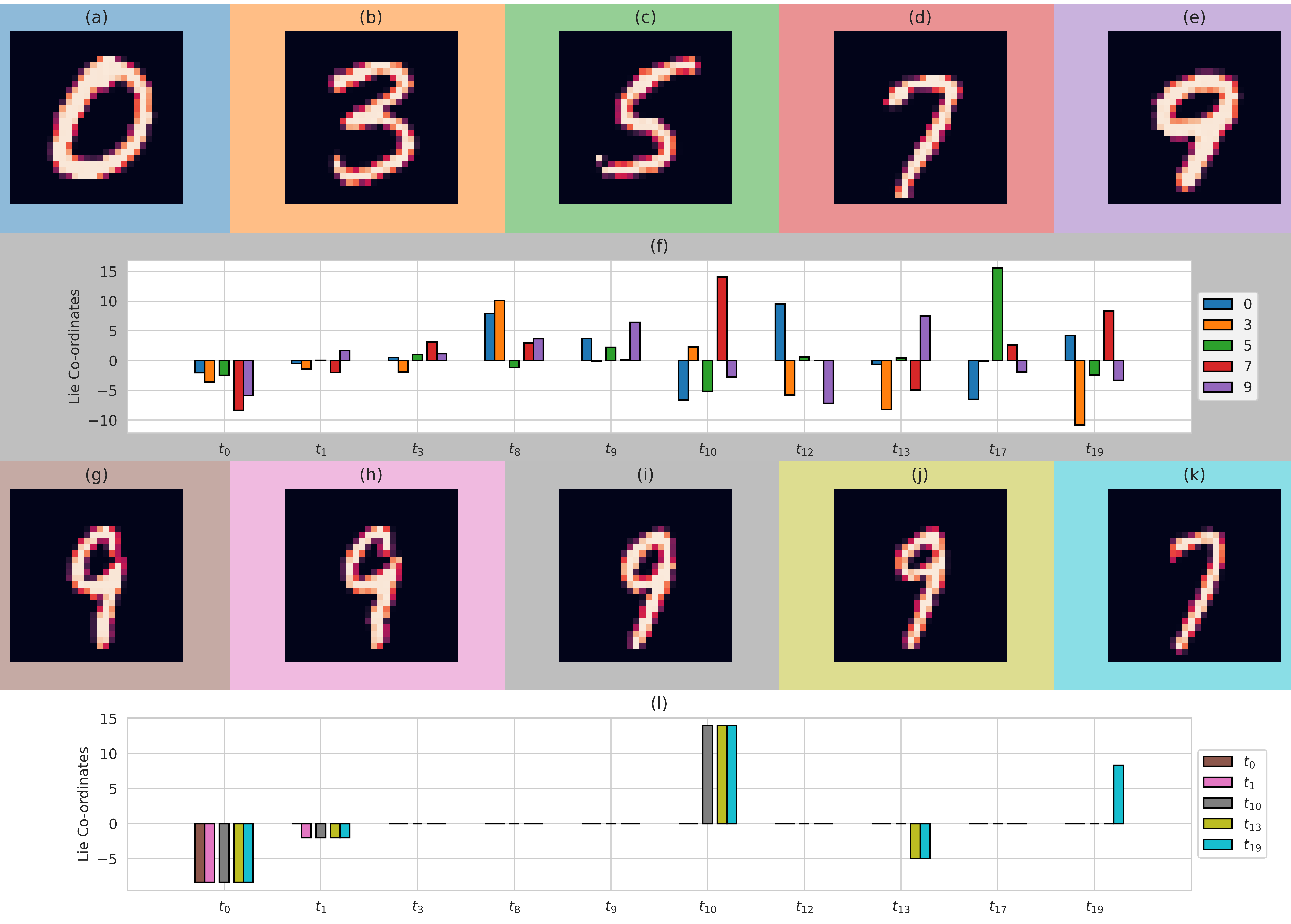}
    \caption{\textbf{Left}: Disentangled conditional generation for dSprites images using Decaf. Each row is obtained by varying the contribution of a specified Lie basis. Lie group coordinate $t_0$ controls the shape between oval, heart and square, $t_1$ controls the scale of the generated image, $t_2$ controls the orientation, $t_3$ controls the position on the X-axis and $t_4$ controls the position on the Y-axis. \textbf{Right}: (\textbf{a)-(e):} MNIST images generated using our model, we provide input labels \{0,3,5,7,9\} to the \emph{Decaf} to obtain corresponding Lie Coordinates and displayed MNIST images \textbf{(f):} We visualize Top 10 Lie coordinates ($t_i$)s that contribute in the generation of the corresponding digits. \textbf{(g)-(k):} We obtain the corresponding values of top five Lie coordinates ($t_i$)s contributed for generating digit '7' shown in \textbf{(l)} and iteratively modify a single Lie coordinate while keeping remaining Lie coordinates to zero. We can clearly observe the formation of digit 7 by controlling  $t_i$s value demonstrating interpretability.
    }
   \label{fig:dsprites}
   \vskip -0.2in
\end{figure}
\paragraph{Results} The summarised results for controllable MNIST generation are summarized in Figure. ~\ref{fig:dsprites} (right) and Table. ~\ref{tab:mnist} respectively.

Table. ~\ref{tab:mnist} supports our claims and shows that our model outperforms the state-of-the-art models for conditional generation including prominent VAE-based FactorVAE, BetaVAE and StyleGAN. \emph{Decaf} attains lower FID as compared to most of the state-of-the-art conditional generative models. The interpretability and controlled conditional generation for MNIST is clearly depicted in the Figure. ~\ref{fig:dsprites} (right)

The key results for disentangled representation are shown for the dSprites dataset in Figure. \ref{fig:dsprites}. Each row is obtained by varying the scalar coordinates for the specified basis. The effective disentanglement of the latent space allows us to smoothly manipulate only one of the generative factors such as the shape of the top row, followed by size, orientation, X position and Y position respectively. It is evident from the figure that varying one of the latent factors   \emph{Decaf} can splendidly control the generation. 
While we observe smooth smooth transitions in the four factors (shape, size, X position, Y position), slight variation can be observed in the orientation due to the multi-scale nature of the flow. Additional figures for a controllable generation with different shapes are shown in Appendix \ref{appendix:dsprites_results}.   
 Table \ref{tab:dsrites} reports the quantitative metrics for disentanglement.  \emph{Decaf} achieves a perfect score for classifier-based metrics \textbf{FVM} and \textbf{DCI} as the Lie-group generator matches each generative factor to one of the Lie group basis. \emph{Decaf} also outperforms the other baselines in \textbf{MIG} that measure how much mutual information about a generative factor is encoded in a single latent factor. \textbf{SAP score} is also comparable for   \emph{Decaf} with the other baselines.

   \emph{Decaf} learns meaningful disentangled representation by matching generative factors with a Lie group basis. This allows for controllable generation while beating existing baselines on 3 out of 4 metrics for disentanglement.

\begin{figure}[ht]
\begin{minipage}[b]{.45\textwidth}
\centering
\includegraphics[width=1\textwidth]{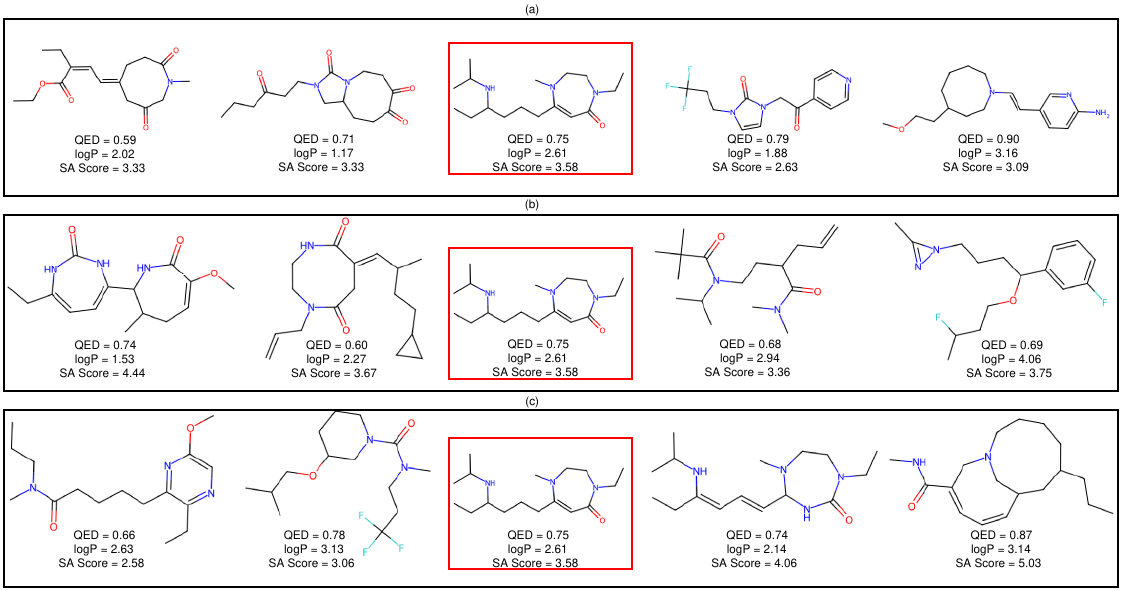}
\caption{We perform controllable generation of molecules using latent traversal in Lie group space by fixing the input noise and only varying a single property at a time. The molecule highlighted with the red bounding box is the starting point. We vary only \textbf{Top row:} QED \textbf{ Middle row:}  logP and \textbf{Bottom row:}  SA Score while keeping the other two constant. The visualized molecules demonstrate traversal across the full range of each property while the other two show minimal variations due to the influence of the coupled component.} \label{fig:mol_disentangled_panel}
\end{minipage}
\hfill
\begin{minipage}[b]{.45\textwidth}
\centering
\includegraphics[width=1\textwidth,height=4.5cm]{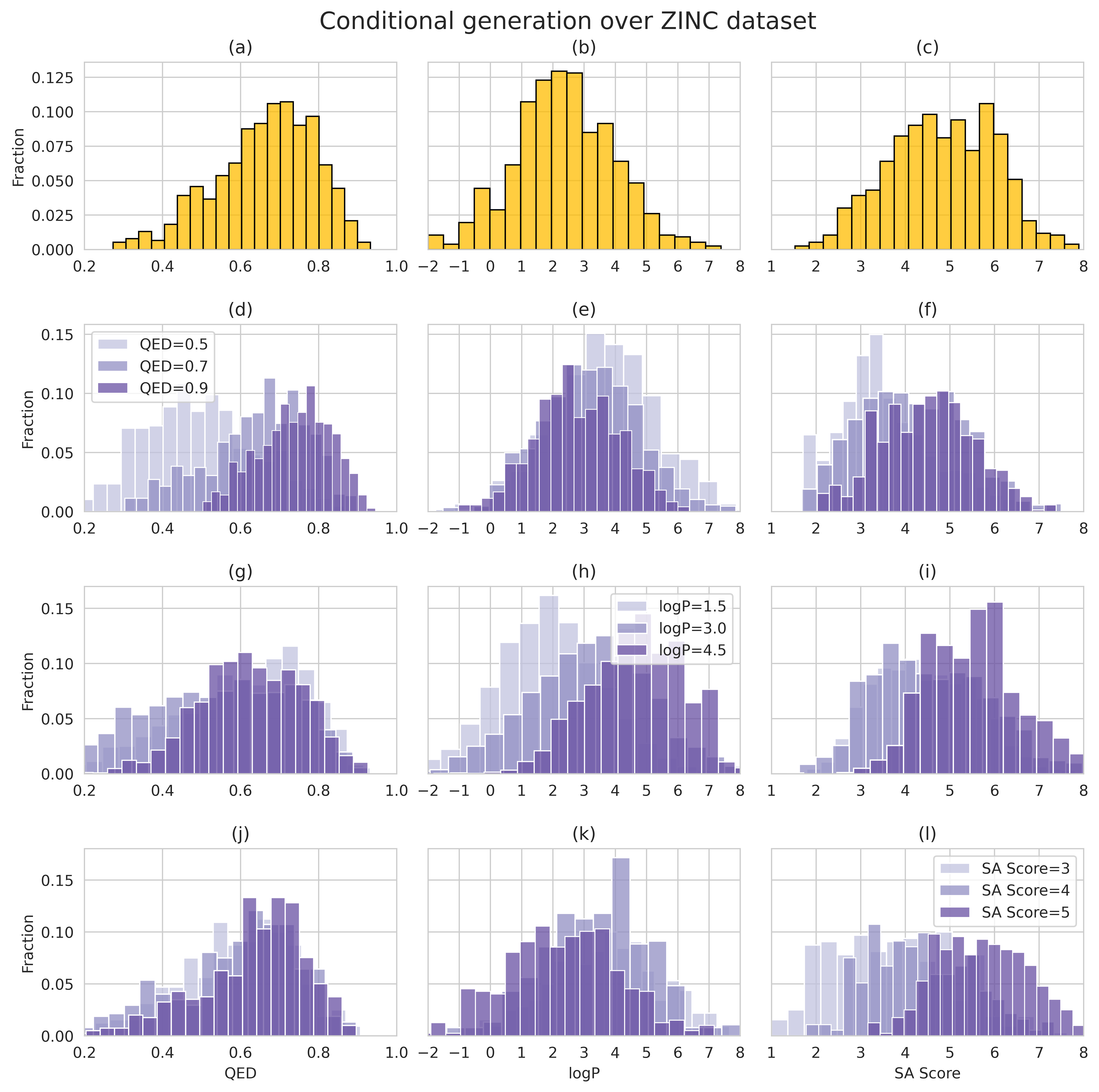}
\caption{Conditional generation with ZINC dataset. (a)-(c) Unconditional generation. (d)-(f) QED target values $\{ 0.5, 0.7, 0.9\}$. (g)-(i) logP target values $\{ 1.5, 3.0, 4.5\}$. (j)-(l) SA Score target values $\{ 3, 4, 5\}$. For each row, non-target properties are randomly sampled to span the range in training data, Decaf successfully shifts the distribution of target property while keeping the other properties similar to unconditional generation.}
    \label{fig:zinc_lie_gen}
\end{minipage}
\end{figure}
\vspace{-0.2in}

\subsection{Molecular generation}
\paragraph{Datasets}
We utilize three common datasets for molecular graph generation to evaluate the performance of   \emph{Decaf} on controllable molecule generation. QM9 dataset \cite{ramakrishnan2014quantum} contains 133k molecules. ZINC dataset \cite{irwin2012zinc} is a set of 250k molecules with at most 38 heavy atoms and 9 atom types. MOSES dataset \cite{10.3389/fphar.2020.565644} contains 1.9M molecules with at most 27 heavy atoms and 7 atom types. 
We use RdKit \cite{landrum_2015_10398} to generate the Quality of Estimated Drug-likeness (QED) and log of the partition coefficient (logP) for all the generated molecules. Synthetic Accessibility score (SA score) \cite{ertl2009estimation} is a measure of how easy a molecule is to synthesize varying between 1 (easy to synthesize) to 10 (difficult to synthesize). We use the code provided in MoFlow \cite{zang2020moflow} to generate the SA Score for all the molecules. QED, logP and SA score are used as labels for conditional generation in \emph{Decaf}. 
\paragraph{Model} QM9 dataset is generated by a variant of Flow++ architecture that uses graph convolution layers to handle graph inputs. For the ZINC and Moses dataset, we use the pre-trained Categorical NF model \cite{lippe2020categorical} which is a state-of-art flow-based generative model for discrete molecular graphs. 
Critic and property predictor share network architecture with the image models described in Section.~\ref{para:img_model}  with different hyper-parameters. Detailed model architecture and hyper-parameters are presented in Appendix.~\ref{appendix:molecular_arch}.
\paragraph{Results} Figure.~\ref{fig:zinc_lie_gen} demonstrate the results of controllable generation of molecules on  \emph{Decaf} model trained on ZINC. Results on QM9 and Moses are shown in Appendix~\ref{appendix:qm9_results}.  
\emph{Decaf} successfully shifts the distribution of the selected property to be centered around the target value. Conditionally generating logP performs well for all three target values. The conditional Lie generator successfully learns a map between input properties and molecular flow latent space. The distributions of other properties show minimal shifts demonstrating the influence of disentangled and coupled blocks. We also report the mean and variance of molecular properties generated with controllable generation using  \emph{Decaf} in Table \ref{tab:conditional_generation}. \emph{Decaf} is competitive with SSVAE \cite{kang2018conditional} baseline for most of the target conditions.
In Appendix~\ref{appendix:qm9_results} consists impressinve results for conditional generation of QM9 dataset.

We compare against existing generative molecular graph methods for property maximization. JT-VAE \cite{jin2018junction} trains a property predictor for each property and perform gradient ascent to find latent vectors with high scores. GCPN \cite{you2018graph} and GraphAF\cite{shi2020graphaf} are auto-regressive baselines that finetunes the generative model with reinforcement learning to bias towards high scores of a single target property. Guided Diffusion \cite{10447350} train a graph diffusion method along with classifier guidance for conditional generation only for QED. The proposed method \emph{Decaf} can successfully generates high scores of targeted property in one shot by simply providing the targeted property to the generator. The auto-regressive baselines may outperform Decaf on penalized logP generation. As they explicitly maximize single targeted property while ignoring the other properties.
On the other hand we can jointly provide the target labels for more than one property without then need of further finetuning. We also visualize the top 6 molecules for both QED and penalized logP in Figure \ref{fig:unconst_plogp_mols}, \ref{fig:unconst_qed_mols} in Appendix \ref{appendix:qm9_results}.

\begin{table}[hb]
\centering
\begin{tabular}{@{}lllllll@{}}
\toprule
Method           & \multicolumn{3}{l}{Penalized logP} & \multicolumn{3}{l}{QED} \\ \midrule
                 & 1st        & 2nd       & 3rd       & 1st    & 2nd    & 3rd   \\
Zinc(Dataset)    & 4.52       & 4.3       & 4.23      & 0.948  & 0.948  & 0.948 \\ \midrule
JT-VAE \cite{jin2018junction}           & 5.3        & 4.93      & 4.49      & 0.925  & 0.911  & 0.910 \\ 
GCPN \cite{you2018graph}            & 7.98       & 7.85      & 7.80      & \textbf{0.948}  & 0.947  & 0.946 \\
GraphAF \cite{shi2020graphaf}         & \textbf{12.23}     & \textbf{11.29}     & \textbf{11.05}     & \textbf{0.948}  & \textbf{0.948}  & 0.947 \\
Guided Diffusion \cite{10447350} & -          & -         & -         & \textbf{0.948}  & \textbf{0.948}  & \textbf{0.948} \\ \midrule
Decaf(Ours)      & 7.06       & 7.00      & 6.78      & \textbf{0.948}  & \textbf{0.948}  & \textbf{0.948} \\ \bottomrule
\end{tabular}
\caption{Unconstrained generation of penalized logP and QED for property maximization evaluated by top-3 scores. Our work \emph{Decaf} is competitive with other methods for QED conditioned generation. The other baselines that utilize reinforcement learning to bias the generative process towards the targeted property outperform \emph{Decaf} for penalized logP. As they explicitly maximize single targeted property while ignoring the other properties.}
\label{tab:unconst_gen}
\vspace{-0.2in}
\end{table}

%% file: conclusion.tex
\section{Conclusion}
In this paper, we present a novel flow-based model \emph{Decaf} that generates the data conditionally.
\emph{Decaf} encodes multi-label conditions directly into a latent space that is aligned to a pre-trained unconditional flow. This allows the conditional generation without growing the flows progressively. 
The entanglement of generative factors hinders controlled generation.
Further, to achieve disentanglement in the latent space for better flexibility and interpretability we introduce Lie group based generator which learns disentangled and entangled latent space aligned to the flow latent space.
We validate the effectiveness of our proposed strategy using important drug-design datasets {QM9, ZINC, MOSES} and the images dataset {MNIST,dSprites}. The superior performance of our proposed method is shown by both qualitative and quantitative results.

\section{Impact Statement}
This paper presents work in application of generative machine learning models for drug discovery. Drug discovery is a costly and time consuming process. In silico methods powered by machine learning have potential to significantly reduce the upfront cost of discovery new drugs. However, there is also potential for these methods to utilized to explore chemical spaces with potential for harm. We believe the impact in reducing costs and time for drug discovery outweighs potential risks of this research.

\newpage
\bibliography{example_paper}
\bibliographystyle{iclr2025_conference}

%% file: appendix.tex
\section{Molecular generation with Flows}
The molecule is represented as a graph $\x = (\A,\B)$, where $\A \in \R^{M \times |\mathcal{A}|}$ is the atom tensor and $\B \in \R^{M \times M \times |\mathcal{B}|}$ is the bond adjacency tensor over $M$ nodes. The atom and bond alphabets $\mathcal{A} = \{\emptyset,C,N,O,\ldots\}$ and $\mathcal{B} = \{\emptyset, 1,2,3,\ldots\}$ include the null element $\emptyset$.

The joint distribution over molecular graphs $p(\A, \B)$ can be decomposed as $p(\A, \B) = p(\A | \B) p(\B)$. Both distributions are modelled by independent flows,
\begin{align}
    &\log p(\A, \B) = \log p(\A | \B)  + \log p(\B) \\
    &= \log \N (f_{A | B}(\A, \B)| \0, I) + \sum_{t=1}^{T_A} \log \left| \frac{\partial f_{A | B}(\A, \B)}{\partial \A} \right| \\ \nonumber
    &\quad + \log \N (f_B (\B) | \0, I) + \sum_{t=1}^{T_B} \log \left| \frac{\partial f_{B}(\B)}{\partial \B} \right|
\end{align}
where $f_{A|B}$ is a conditional flow with graph coupling layers and $f_B$ is a flow with convolution layers

\section{Image generation} \label{appendix:dsprites_results}
\begin{figure*}[b]
    \centering
    \includegraphics[width=\linewidth]{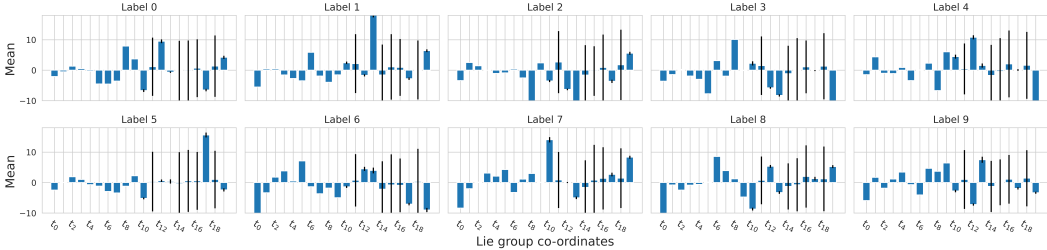}
    \caption{Coordinates $t_i$s distribution for the corresponding  Lie group basis for MNIST images. We report the mean and variance of each coordinate over 10,000 noise vectors per label. The coordinates with no variance control the shape of the digit for an appropriate label and coordinates with high variance correspond to the style variations for each digit such as line width, rotation etc.}
    \label{fig:mnist_lie_hist}
\end{figure*}

 \begin{figure}[ht]
\vskip 0.2in
    \centering
    \includegraphics[width=\linewidth,height=4cm]{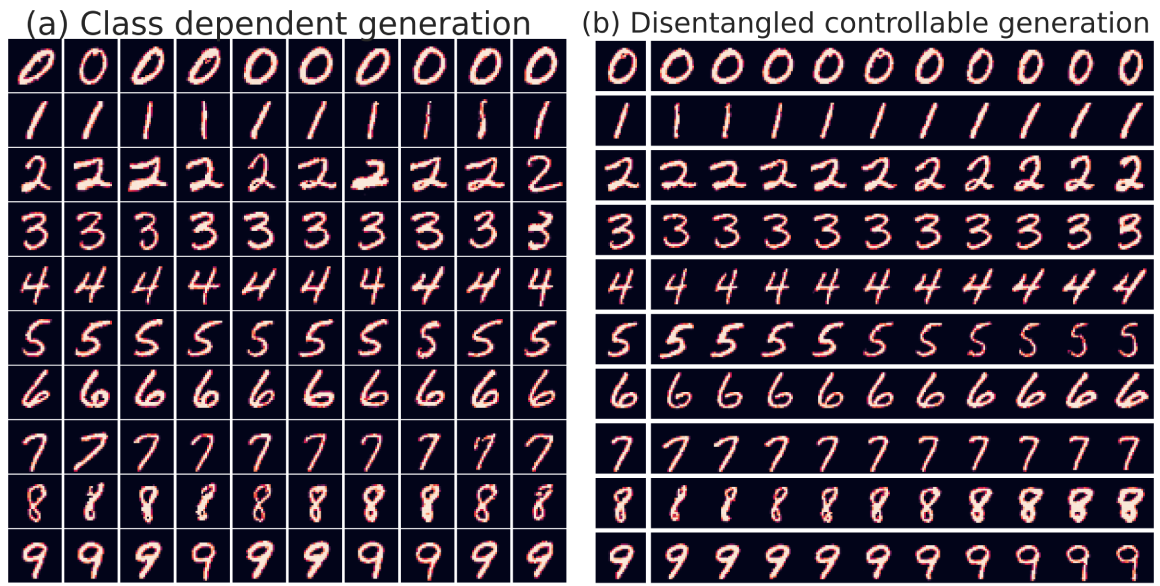}
    \caption{\textbf{(a)} Conditional generation of MNIST images with Lie Generator for different classes \textbf{(b)} Disentangled image manipulation by varying a single Lie group basis for each class. We start from the first image in each row and gradually vary a single Lie coordinate. The disentangled representation allows us to modify independent style elements like the width of '0' and the rotation of '1'.}
    \label{fig:mnist}
    \vskip  -0.1in
\end{figure}
\begin{figure}[b]
    \centering
    \includegraphics[width=0.45\linewidth]{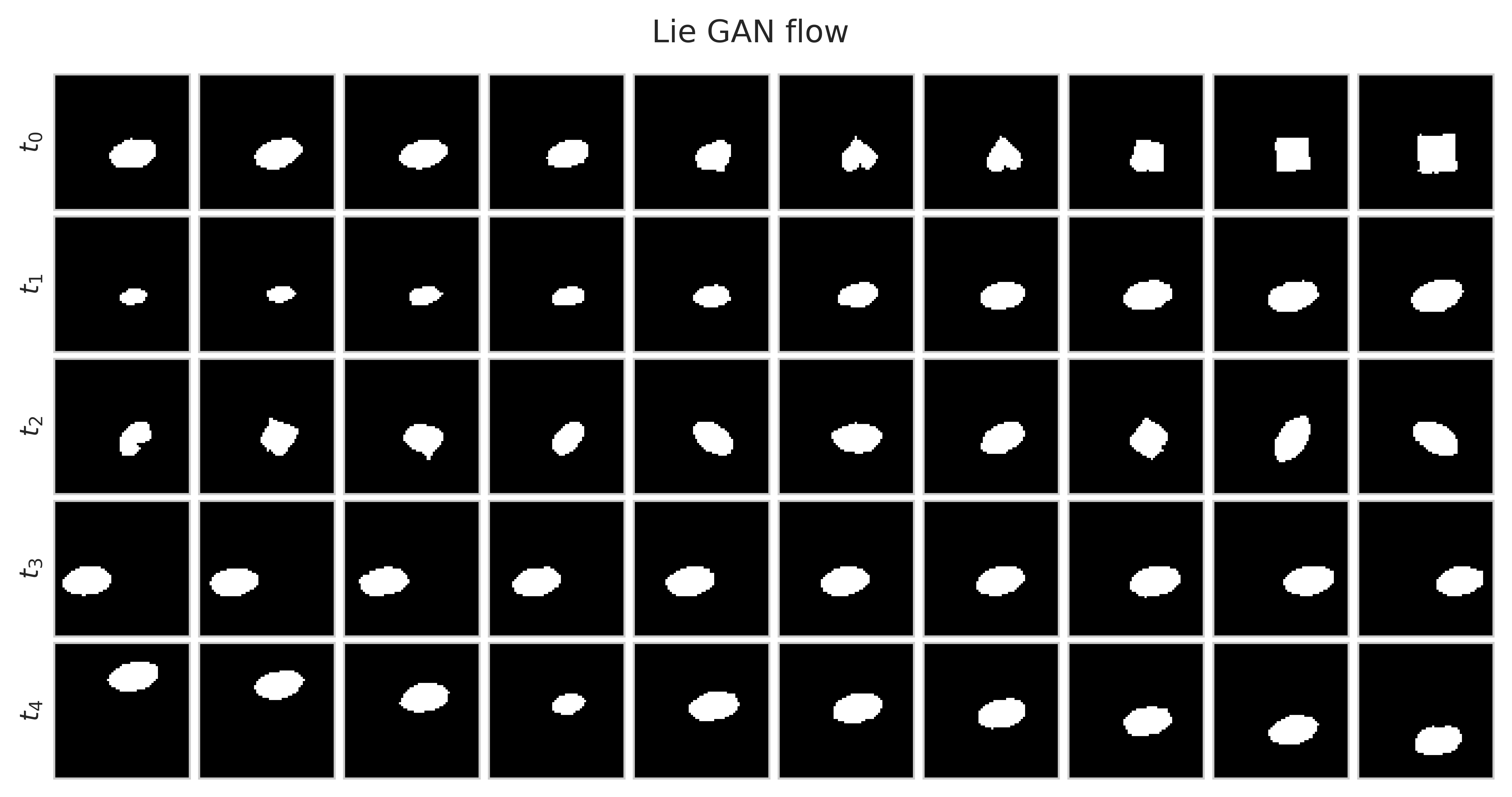}
    \includegraphics[width=0.45\linewidth]{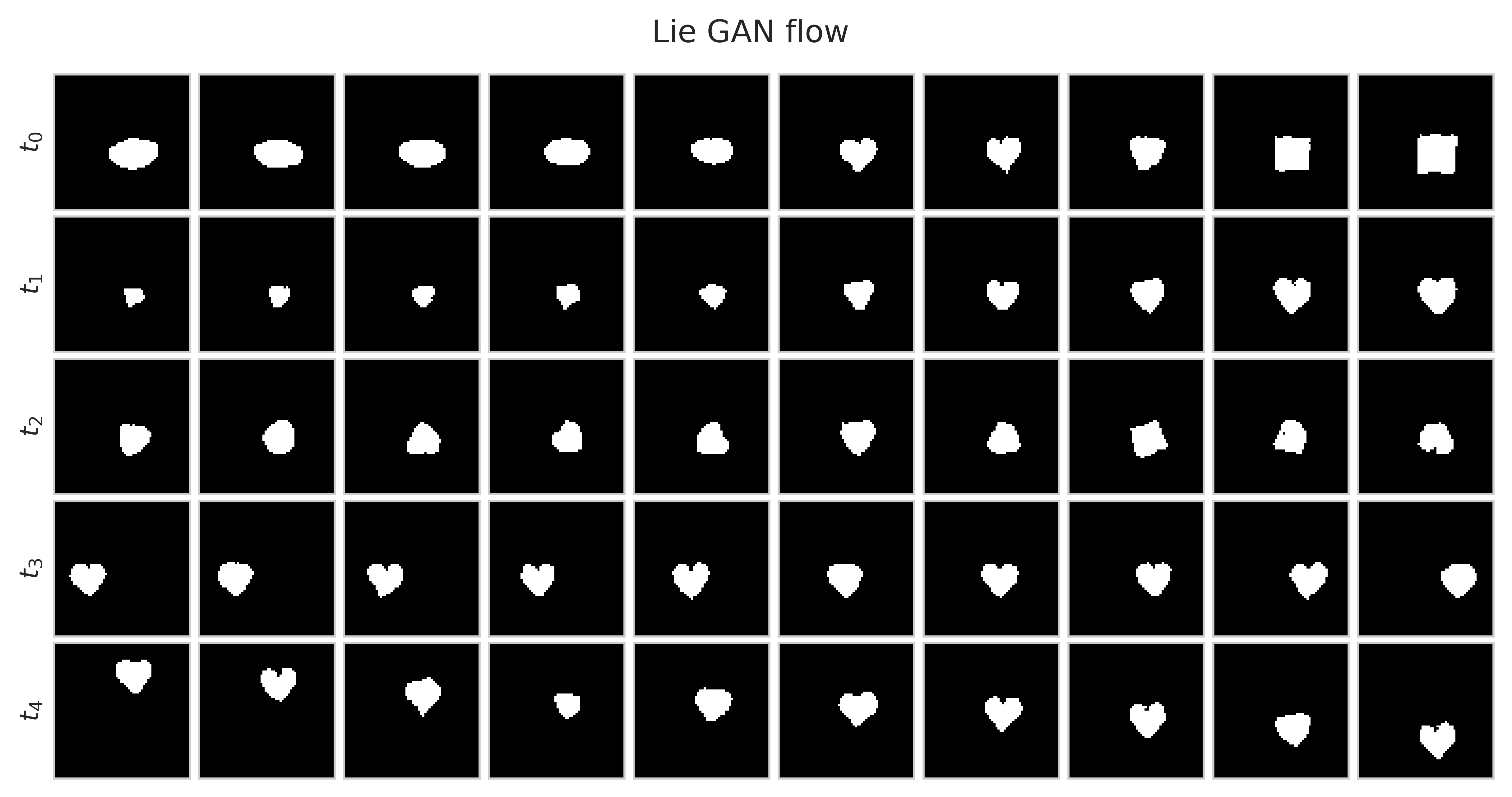}
    \caption{Disentangled conditional generation with other shapes as default vector. }
    \label{fig:lie_dsprites_shapes}
\end{figure}
\begin{figure}
    \centering
    \includegraphics[width=\linewidth]{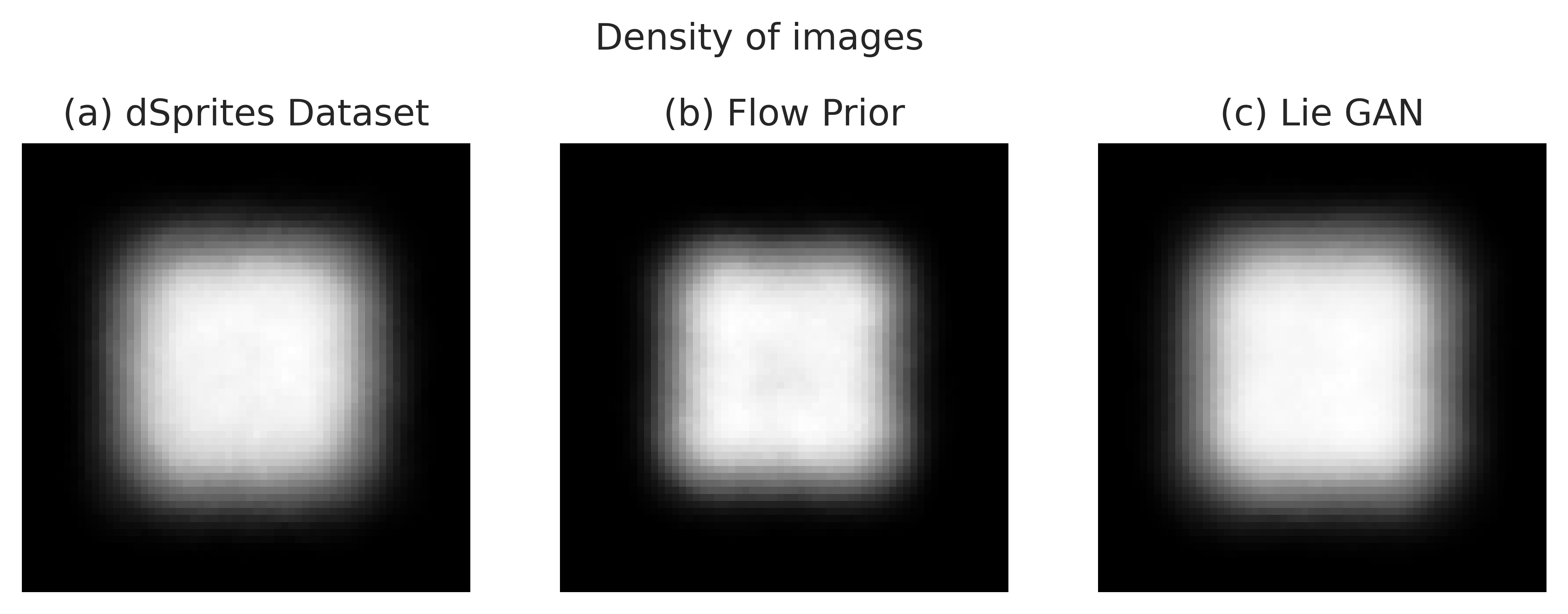}
    \caption{Sample density from (a) dSprites dataset (b) prior of normalizing flow (c) randomly generated from Lie GAN.}
    \label{fig:lie_dsprites_density}
\end{figure}
Decaf also allows for interpretable generation using the scalar Lie coordinates to control different aspects of image generation. Figure \ref{fig:mnist_lie_hist} shows the histogram for Lie coordinates averaged over 1000 samples per label. The Lie basis with small variance contributes to defining the general shape for each class. The basis with high variance is associated with style for each label and varying them also us to control underlying generative factors such as width, rotation etc for different labels.

We perform latent traversal over all 5 Lie basis for dSprites dataset using different initial values corresponding to different shapes in Fig \ref{fig:lie_dsprites_shapes}. We observe similar results on controllable generation for all three shapes. The multi-scale nature of the normalizing flow used adds some variance to the shape as half of the pixel values are obtained from the prior. 

\begin{figure}
    \centering
    \includegraphics[width=\linewidth]{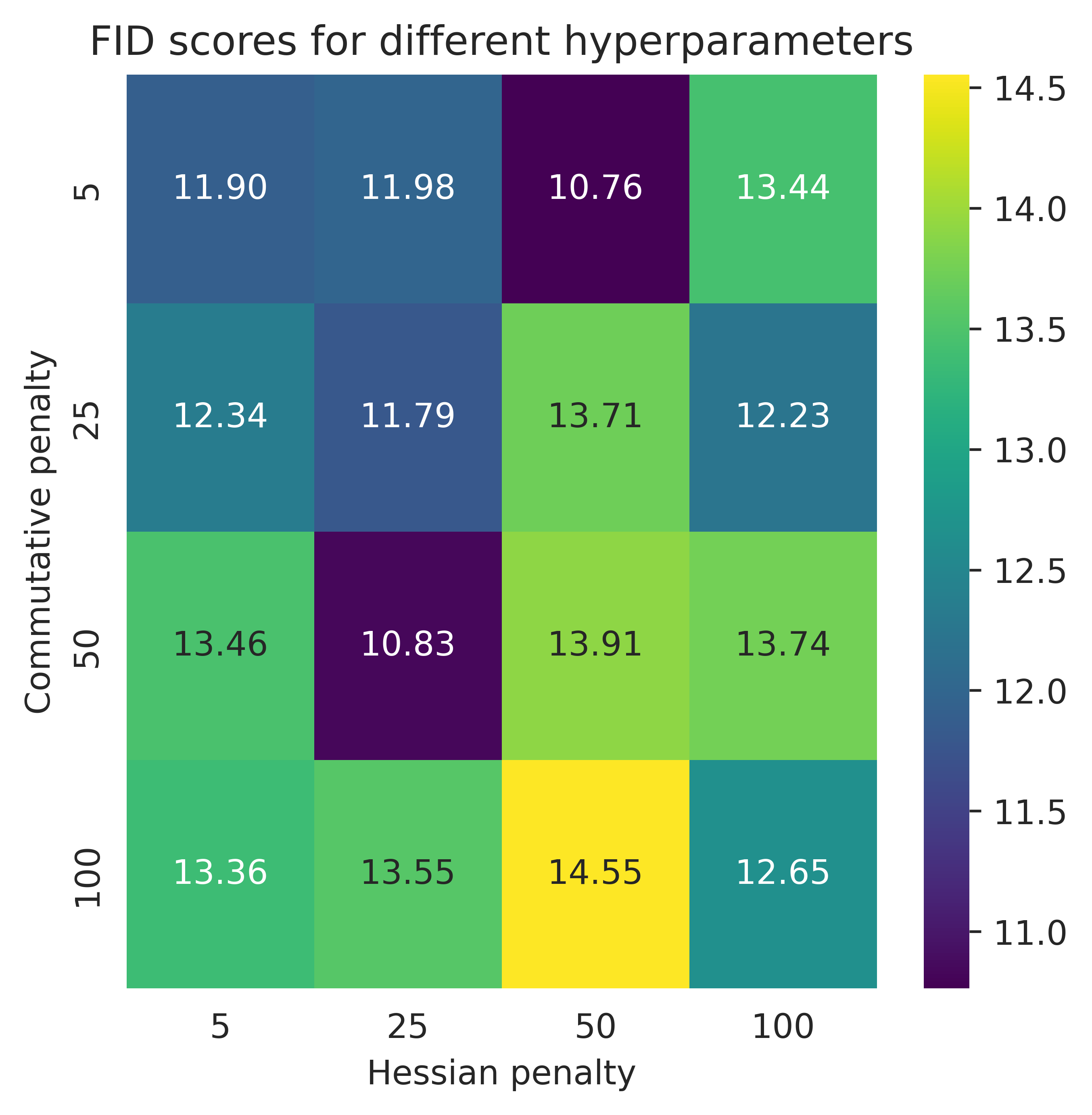}
    \caption{Caption}
    \label{fig:enter-label}
\end{figure}
\section{Image generation models} \label{appendix:img_gen_model}
The affine coupling layers used for imaged generation use Gated Convolution networks. We use variational dequantization to map discrete pixel values to a continuous domain. Following the SOTA architecture, we use Actnorm and 1x1 convolution prior to the affine coupling layer for each transformation.

The code for the Affine Coupling layers is,
\begin{verbatim}{python}
class ConcatELU(nn.Module):
    """
    Activation function that applies ELU in both direction (inverted and plain).
    Allows non-linearity while providing strong gradients for any input (important for final convolution)
    """

    def forward(self, x):
        return torch.cat([F.elu(x), F.elu(-x)], dim=1)

class LayerNormChannels(nn.Module):

    def __init__(self, c_in, eps=1e-5):
        """
        This module applies layer norm across channels in an image.
        Inputs:
            c_in - Number of channels of the input
            eps - Small constant to stabilize std
        """
        super().__init__()
        self.gamma = nn.Parameter(torch.ones(1, c_in, 1, 1))
        self.beta = nn.Parameter(torch.zeros(1, c_in, 1, 1))
        self.eps = eps

    def forward(self, x):
        mean = x.mean(dim=1, keepdim=True)
        var = x.var(dim=1, unbiased=False, keepdim=True)
        y = (x - mean) / torch.sqrt(var + self.eps)
        y = y * self.gamma + self.beta
        return y

class GatedConv(nn.Module):

    def __init__(self, c_in, c_hidden):
        """
        This module applies a two-layer convolutional ResNet block with input gate
        Inputs:
            c_in - Number of channels of the input
            c_hidden - Number of hidden dimensions we want to model (usually similar to c_in)
        """
        super().__init__()
        self.net = nn.Sequential(
            ConcatELU(),
            nn.Conv2d(2*c_in, c_hidden, kernel_size=3, padding=1),
            ConcatELU(),
            nn.Conv2d(2*c_hidden, 2*c_in, kernel_size=1)
        )

    def forward(self, x):
        out = self.net(x)
        val, gate = out.chunk(2, dim=1)
        return x + val * torch.sigmoid(gate)

class GatedConvNet(nn.Module):

    def __init__(self, c_in, c_hidden=32, c_out=-1, num_layers=3):
        """
        Module that summarizes the previous blocks to a full convolutional neural network.
        Inputs:
            c_in - Number of input channels
            c_hidden - Number of hidden dimensions to use within the network
            c_out - Number of output channels. If -1, 2 times the input channels are used (affine coupling)
            num_layers - Number of gated ResNet blocks to apply
        """
        super().__init__()
        c_out = c_out if c_out > 0 else 2 * c_in
        layers = []
        layers += [nn.Conv2d(c_in, c_hidden, kernel_size=3, padding=1)]
        for layer_index in range(num_layers):
            layers += [GatedConv(c_hidden, c_hidden),
                       LayerNormChannels(c_hidden)]
        layers += [ConcatELU(),
                   nn.Conv2d(2*c_hidden, c_out, kernel_size=3, padding=1)]
        self.nn = nn.Sequential(*layers)

        self.nn[-1].weight.data.zero_()
        self.nn[-1].bias.data.zero_()

    def forward(self, x):
        return self.nn(x)
\end{verbatim}
Invertible transformation is performed using affine coupling where the input $\mathbf{z}^i$ at step $i$ is split in two halves $\mathbf{z}_{1:D/2}$ and $\mathbf{z}_{D/2:D}$. The forward transformation is defined over half the latent dimensions as,
\begin{align}
    \mathbf{z}_{D/2:D}^{i+1} &= ( \mathbf{z}_{D/2:D}^{i} - s(\mathbf{z}_{1:D/2}^{i}) / t(\mathbf{z}_{1:D/2}^{i}) \\
    \mathbf{z}_{1:D/2}^{i+1} &= \mathbf{z}_{1:D/2}^{i+1}
\end{align}
where $s$ and $t$ are obtrained by the Gated Convolution network defined above. Multi-scale architectures use Squeeze and Split layers to reduce the dimension of latent dimensions that are transformed. 
\begin{verbatim}{python}
    class SqueezeFlow(nn.Module):

    def forward(self, z, ldj, reverse=False):
        B, C, H, W = z.shape
        if not reverse:
            # Forward direction: H x W x C => H/2 x W/2 x 4C
            z = z.reshape(B, C, H//2, 2, W//2, 2)
            z = z.permute(0, 1, 3, 5, 2, 4)
            z = z.reshape(B, 4*C, H//2, W//2)
        else:
            # Reverse direction: H/2 x W/2 x 4C => H x W x C
            z = z.reshape(B, C//4, 2, 2, H, W)
            z = z.permute(0, 1, 4, 2, 5, 3)
            z = z.reshape(B, C//4, H*2, W*2)
        return z, ldj

    class SplitFlow(nn.Module):

    def __init__(self):
        super().__init__()
        self.prior = torch.distributions.normal.Normal(loc=0.0, scale=1.0)

    def forward(self, z, ldj, reverse=False):
        if not reverse:
            z, z_split = z.chunk(2, dim=1)
            ldj += self.prior.log_prob(z_split).sum(dim=[1,2,3])
        else:
            z_split = self.prior.sample(sample_shape=z.shape).to(device)
            z = torch.cat([z, z_split], dim=1)
            ldj -= self.prior.log_prob(z_split).sum(dim=[1,2,3])
        return z, ldj
\end{verbatim}
The full normalizing flow for MNIST and dSprites is defined in Table.
\begin{table}[]
\caption{Detailed normalizing flow architecture for MNIST and dSprites dataset.}
\begin{tabular}{@{}lll@{}}
\toprule
\multicolumn{3}{c}{MNIST Flow} \\
\midrule
Layer & Output Size & Description \\ \midrule
Input & $1 \times 28 \times 28$ & MNIST Image\\
Variational Dequantization& $1 \times 28 \times 28$ & Invertible transformation from discrete pixel space to continuous space \\
\cmidrule{1-2}
Actnorm & $1 \times 28 \times 28$ &             \\ 
1$\times$1 Invertible convolution  & $1 \times 28 \times 28$ & One block of invertible transformation. Repeated 16 times. \\ 
Affine coupling layer & $1 \times 28 \times 28$ &             \\ 
\cmidrule{1-2}
Squeeze layer & $4 \times 14 \times 14$ & Transformation to increase channels \\ 
Split layer & $2 \times 14 \times 14$ & Transformation to reduce dimensionality \\
\cmidrule{1-2}
Actnorm & $2 \times 14 \times 14$ &             \\ 
1$\times$1 Invertible convolution  & $2 \times 14 \times 14$ & 16 blocks. \\ 
Affine coupling layer & $2 \times 14 \times 14$  &             \\ 
\cmidrule{1-2}
Squeeze layer & $8 \times 7 \times 7$ & Transformation to increase channels \\ 
Split layer & $4 \times 7 \times 7$ & Transformation to reduce dimensionality \\
\cmidrule{1-2}
Actnorm & $4 \times 7 \times 7$ &             \\ 
1$\times$1 Invertible convolution  & $4 \times 7 \times 7$ & 16 blocks. \\ 
Affine coupling layer & $4 \times 7 \times 7$  &             \\ 
\midrule
\multicolumn{3}{c}{dSprites Flow} \\
\midrule
Layer & Output Size & Description \\ \midrule
Input & $1 \times 64 \times 64$ & dSprites Image\\
Variational Dequantization& $1 \times 64 \times 64$ & Invertible transformation from discrete pixel space to continuous space \\
\cmidrule{1-2}
Actnorm & $1 \times 64 \times 64$ &             \\ 
1$\times$1 Invertible convolution  & $1 \times 64 \times 64$ & One block of invertible transformation. Repeated 8 times. \\ 
Affine coupling layer & $1 \times 64 \times 64$ &             \\ 
\cmidrule{1-2}
Squeeze layer & $4 \times 32 \times 32$ & Transformation to increase channels \\ 
Split layer & $2 \times 32 \times 32$ & Transformation to reduce dimensionality \\
\cmidrule{1-2}
Actnorm & $2 \times 32 \times 32$ &             \\ 
1$\times$1 Invertible convolution  & $2 \times 32 \times 32$ & 8 blocks. \\ 
Affine coupling layer & $2 \times 32 \times 32$  &             \\ 
\cmidrule{1-2}
Squeeze layer & $8 \times 16 \times 16$ & Transformation to increase channels \\ 
Split layer & $4 \times 16 \times 16$ & Transformation to reduce dimensionality \\
\cmidrule{1-2}
Actnorm & $4 \times 16 \times 16$ &             \\ 
1$\times$1 Invertible convolution  & $4 \times 16 \times 16$ & 2 blocks. \\ 
Affine coupling layer & $4 \times 16 \times 16$  &             \\ 
\bottomrule
\end{tabular}
\end{table}
MNIST Flow has uses affine coupling layers with hidden dim of 128 and dSprites flow has hidden dimension of 32. We train both flows using Adam optimizer with learning rate $1e-3$ for 100 epochs.

\subsection{Disentanglement matrices} \label{appendix:dis_metrics}
We rely on various popular metrics such as FactorVAE Metric(FVM) \cite{kim2018disentangling}, DCI \cite{eastwood2018a}, Seperated Attribute Predictability (SAP) score \cite{kumar2017variational} and Mutual Information Gap (MIG) \cite{chen2018isolating}. 
To predict the corresponding generative factor, \textbf{FactorVAE} Metric employs a linear classifier over the variation of latent factors. 
\textbf{DCI}, on the other hand, calculates an important score for every pair by using independent random forest models to predict generative elements from the latent codes. The weighted sum of entropy over the important factor for each latent factor that is derived from the regressors is the disentanglement score. \textbf{MIG} assesses the mutual information among all latent and generative components without the need for classifiers. The difference between each generative factor's top two MI values, normalised by entropy, yields the score. The given average score is the sum of all the factors. \textbf{SAP} similarly measures the correlation instead of MI in the case of continuous factors or a decision tree for discrete factors. The difference between scores of the top 2 latent factors for each generative factor is considered.

\section{Molecular generation with Flows}
The molecule is represented as a graph $\x = (\A,\B)$, where $\A \in \R^{M \times |\mathcal{A}|}$ is the atom tensor and $\B \in \R^{M \times M \times |\mathcal{B}|}$ is the bond adjacency tensor over $M$ nodes. The atom and bond alphabets $\mathcal{A} = \{\emptyset,C,N,O,\ldots\}$ and $\mathcal{B} = \{\emptyset, 1,2,3,\ldots\}$ include the null element $\emptyset$.

The joint distribution over molecular graphs $p(\A, \B)$ can be decomposed as $p(\A, \B) = p(\A | \B) p(\B)$. Both distributions are modelled by independent flows,
\begin{align}
    &\log p(\A, \B) = \log p(\A | \B)  + \log p(\B) \\
    &= \log \N (f_{A | B}(\A, \B)| \0, I) + \sum_{t=1}^{T_A} \log \left| \frac{\partial f_{A | B}(\A, \B)}{\partial \A} \right| \\ \nonumber
    &\quad + \log \N (f_B (\B) | \0, I) + \sum_{t=1}^{T_B} \log \left| \frac{\partial f_{B}(\B)}{\partial \B} \right|
\end{align}
where $f_{A|B}$ is a conditional flow with graph coupling layers and $f_B$ is a flow with convolution layers

\section{QM9} \label{appendix:qm9_results}
\begin{figure}
    \centering
    \includegraphics[width=\linewidth]{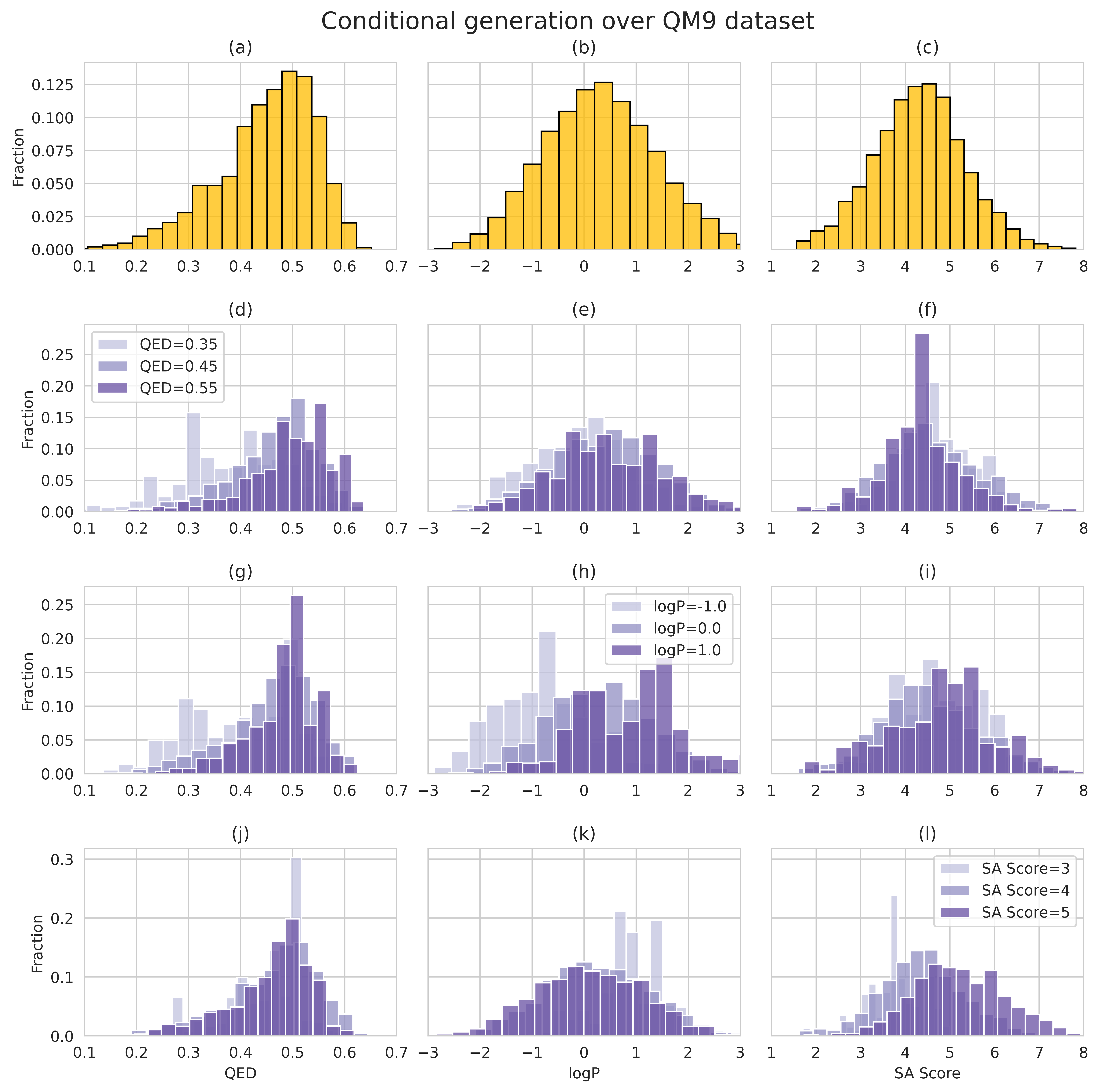}
    \caption{Conditional generation with QM9}
    \label{fig:qm9_lie_gen}
\end{figure}
Controllable generation using the QM9 dataset is show in Fig \ref{fig:qm9_lie_gen}. The QED has target values of $\{0.35, 0.45, 0.55\}$, logP has $\{-1, 0, 1\}$ and SA Score has target $\{3, 4, 5\}$.

\section{Molecular models} \label{appendix:molecular_arch}
\begin{figure}
    \centering\includegraphics[width=0.9\linewidth,height=6.5cm]{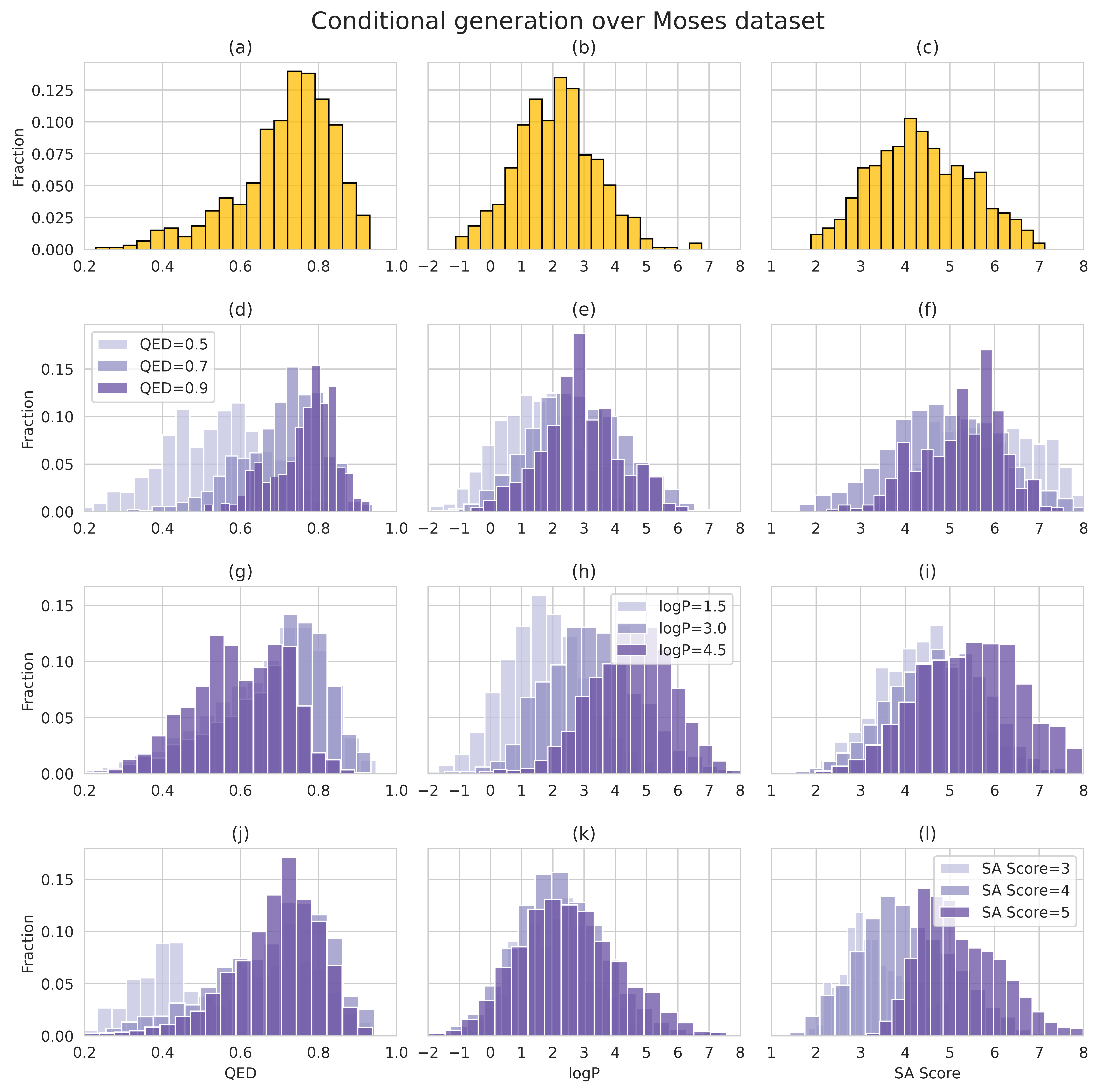}
    \caption{Conditional generation with Moses dataset. (a)-(c) Unconditional generation. (d)-(f) QED target values $\{ 0.5, 0.7, 0.9\}$. (g)-(i) logP target values $\{ 1.5, 3.0, 4.5\}$. (j)-(l) SA Score target values $\{ 3, 4, 5\}$. Decaf successfully shifts the distribution of target property while keeping the other properties similar to unconditional generation.}
    \label{fig:moses_lie_gen}
\end{figure}
\begin{table}[]
\caption{Validity, Uniqueness and Novelty of various graph generative models. We report validity without using any post-processing methods.} \label{tab:vun}
\begin{tabular}{@{}llll@{}}
\toprule
\multicolumn{4}{c}{ZINC dataset} \\
\midrule
Model & Validity & Uniqueness & Novelty \\ \midrule
MRNN & 65\% & 100\% & 100\%   \\
GCPN & 20\% & 100\% & 100\%   \\
MoFlow & 82\% & 100\% & 100\% \\ 
GraphAF & 68\% & 99\% & 100\% \\
GraphCNF & 83\% & 100\% & 100\% \\
Decaf (ours) & 78\% & 100\% & 100\% \\
\midrule
\multicolumn{4}{c}{Moses dataset} \\
\midrule
GraphAF & 71\% & 99\% & 100\% \\
GraphCNF & 82\% & 100\% & 100\% \\
Decaf (ours) & 81\% & 100\% & 100\% \\
\bottomrule
\end{tabular}
\end{table}
\begin{table}[]
\caption{Hyper parameters for Decaf trained on QM9 model} \label{tab:qm9_hp}
\begin{tabular}{@{}lll@{}}
\toprule
Hyper parameter & Value & Description \\ \midrule
$\lambda_{GP}$& 20 & Weight of gradient penalty \\
$\lambda_{P}$& 0.1 & Weight of property predictor \\
$\lambda_{F}$& 1 & Weight of flow loss \\
$\lambda_{H}$& 10 & Weight of Hessian penalty \\ 
$\lambda_{C}$& 20 & Weight of Commutative penalty \\ 
Critic depth& 1 & Number of Residual blocks in Critic \\ 
Prop depth& 2 & Number of Residual blocks in Property prediction network \\ 
T & 8 & Number of Lie basis \\ 
$|\lA|$ & $7 \times 7$ & Dimensionality of Lie basis \\
$|\epsilon|$ & 10 & Dimensionality of noise vector \\
\bottomrule
\end{tabular}
\end{table}
\begin{table}[]
\caption{Hyper parameters for Decaf trained on ZINC model} \label{tab:zinc_hp}
\begin{tabular}{@{}lll@{}}
\toprule
Hyper parameter & Value & Description \\ \midrule
$\lambda_{GP}$& 10 & Weight of gradient penalty \\
$\lambda_{P}$& 0.5 & Weight of property predictor \\
$\lambda_{F}$& 1 & Weight of flow loss \\
$\lambda_{H}$& 5 & Weight of Hessian penalty \\ 
$\lambda_{C}$& 20 & Weight of Commutative penalty \\ 
Critic depth& 3 & Number of Residual blocks in Critic \\ 
Prop depth& 5 & Number of Residual blocks in Property prediction network \\ 
T & 6 & Number of Lie basis \\ 
$|\lA|$ & $40 \times 40$ & Dimensionality of Lie basis \\
$|\epsilon|$ & 20 & Dimensionality of noise vector \\
\bottomrule
\end{tabular}
\end{table}
\begin{table}[]
\caption{Hyper parameters for Decaf trained on Moses model} \label{tab:moses_hp}
\begin{tabular}{@{}lll@{}}
\toprule
Hyper parameter & Value & Description \\ \midrule
$\lambda_{GP}$& 10 & Weight of gradient penalty \\
$\lambda_{P}$& 0.5 & Weight of property predictor \\
$\lambda_{F}$& 1 & Weight of flow loss \\
$\lambda_{H}$& 5 & Weight of Hessian penalty \\ 
$\lambda_{C}$& 20 & Weight of Commutative penalty \\ 
Critic depth& 3 & Number of Residual blocks in Critic \\ 
Prop depth& 5 & Number of Residual blocks in Property prediction network \\ 
T & 6 & Number of Lie basis \\ 
$|\lA|$ & $29 \times 29$ & Dimensionality of Lie basis \\
$|\epsilon|$ & 20 & Dimensionality of noise vector \\
\bottomrule
\end{tabular}
\end{table}
We report the Validity, Uniqueness and Novelty of baselines and Decaf in Table \ref{tab:vun}. Decaf obtains comparable scores and can generate valid molecules while allowing for controllable generation. We also report the hyper parameters used in Decaf for QM9, Zinc and Moses dataset in Tables \ref{tab:qm9_hp}, \ref{tab:zinc_hp} and \ref{tab:moses_hp}.

\begin{table}[]
\caption{ Comparison of conditional generation results for our model  \emph{Decaf}}
\adjustbox{max width=0.5\textwidth}{%
\begin{tabular}{@{}llllll@{}}
\toprule                                                   
 Model & Condition    & QED             & logP            & SA Score        \\ \midrule
 \multicolumn{5}{c}{ZINC} \\
 \midrule
 ZINC dataset & -           & $0.74 \pm 0.13$ & $2.45 \pm 1.44$ & $3.06 \pm 0.84$ \\
 \cmidrule{2-5}
  & QED=0.5      & $0.53 \pm 0.09$ & $3.29 \pm 1.07$ & - \\
  & QED=0.7      &$0.72 \pm 0.08$&$2.89 \pm 1.09$& - \\
  & QED=0.9      &$0.84 \pm 0.07$&$2.44 \pm 1.08$ - \\ 
SSVAE & logP=1.5     &$0.75 \pm 0.13$&$1.54 \pm 0.30$& - \\ 
  &logP=3.0     &$0.70 \pm 0.15$&$2.98 \pm 0.30 $& - \\ 
  & logP=4.5     &$0.62 \pm 0.15$&$4.35 \pm 0.31$& - \\
\cmidrule{2-5}
  & unconditional & $0.66 \pm 0.13$ & $2.37 \pm 1.60$ & $4.82 \pm 1.18$ \\
  & QED=0.5      & $0.49 \pm 0.14$ & $3.63 \pm 1.65$ & $3.60 \pm 1.13$ \\
  & QED=0.7      &$0.67 \pm 0.13$&$3.22 \pm 1.55$&$4.14 \pm 1.18$\\
 & QED=0.9      &$0.78 \pm 0.09$&$2.77 \pm 1.35$&$4.49 \pm 1.04$\\ 
Decaf & logP=1.5     &$0.60 \pm 0.16$&$1.91 \pm 1.68$&$4.45 \pm 1.12$\\ 
  &logP=3.0     &$0.54 \pm 0.18$&$3.17 \pm 1.91$&$4.57 \pm 1.28$\\ 
  & logP=4.5     &$0.62 \pm 0.13$&$4.68 \pm 1.52$&$5.52 \pm 1.02$\\ 
 & SA Score=3   &$0.59 \pm 0.14$&$3.70 \pm 1.69$&$3.73 \pm 1.36$\\ 
  & SA Score=4   &$0.58 \pm 0.17$ & $3.49 \pm 1.74$ & $4.36 \pm 1.14$\\ 
  & SA Score=5   &$0.63 \pm 0.14$ & $2.34 \pm 1.75$ & $5.71 \pm 1.02$\\ 
 \midrule
\multicolumn{5}{c}{Moses}                                               \\ \midrule
Moses Dataset & - & $0.81 \pm 0.09$ & $2.44 \pm 0.93$ & $2.45 \pm 0.46$ \\
\cmidrule{2-5}
& Unconditional & $0.71 \pm 0.13$ & $2.13 \pm 1.41$ & $4.45 \pm 1.01$ \\
 & QED=0.5      & $0.54 \pm 0.13$ & $1.49 \pm 1.28$ & $5.78 \pm 1.15$ \\
  & QED=0.7     &$0.71 \pm 0.10$ & $2.80 \pm 1.50$ & $4.87 \pm 1.24$\\
  & QED=0.9      &$0.77 \pm 0.08$ & $2.89 \pm 1.24$ & $5.27 \pm 0.93$\\ 
 & logP=1.5     &$0.67 \pm 0.14$ & $1.76 \pm 1.32$ & $4.43 \pm 0.99$\\ 
  \emph{Decaf} & logP=3.0     &$0.69 \pm 0.14$ & $3.26 \pm 1.48$ & $4.73 \pm 1.06$\\ 
  & logP=4.5     &$0.60 \pm 0.12$ & $4.52 \pm 1.31$ & $5.44 \pm 1.17$\\ 
 & SA Score=3   &$0.58 \pm 0.18$ & $2.04 \pm 1.17$ & $3.36 \pm 0.73$\\ 
  & SA Score=4   &$0.68 \pm 0.15$ & $2.23 \pm 1.41$ & $4.02 \pm 1.05$\\ 
  & SA Score=5   &$0.69 \pm 0.12$ & $2.53 \pm 1.61$ & $5.22 \pm 0.90$\\  
\midrule
\multicolumn{5}{c}{QM9}                                               \\ \midrule
QM9 Dataset & - & $0.47 \pm 0.07$ & $0.30 \pm 1.00$ & $4.26 \pm 0.95$ \\
 \cmidrule{2-5}
 & Unconditional & $0.45 \pm 0.10$ & $0.31 \pm 1.05$ & $4.30 \pm 1.01$ \\
 & qed=0.35 & $0.38 \pm 0.10$ & $-0.13 \pm 0.93$ & $4.75 \pm 0.89$ \\
 & qed=0.45 & $0.46 \pm 0.08$ & $0.37 \pm 1.02$ & $4.58 \pm 1.09$ \\
 & qed=0.55 & $0.50 \pm 0.08$ & $0.47 \pm 1.04$ & $4.38 \pm 0.90$ \\
 & logP=-1.0 & $0.39 \pm 0.10$ & $-0.82 \pm 0.89$ & $4.75 \pm 0.97$ \\
Decaf & logP=0.0 & $0.46 \pm 0.09$ & $0.28 \pm 0.99$ & $4.53 \pm 1.11$ \\
 & logP=1.0 & $0.48 \pm 0.07$ & $0.88 \pm 0.97$ & $4.78 \pm 1.19$ \\
 & SA Score=3 & $0.45 \pm 0.07$ & $0.91 \pm 0.67$ & $3.77 \pm 0.61$ \\
 & SA Score=4 & $0.46 \pm 0.09$ & $0.26 \pm 1.00$ & $4.37 \pm 0.95$ \\
 & SA Score=5 & $0.46 \pm 0.08$ & $0.10 \pm 1.04$ & $5.21 \pm 1.01$ \\
\bottomrule
\end{tabular}} \label{tab:conditional_generation}
\end{table}

\begin{table}[]
\begin{tabular}{@{}ccccc@{}}
\toprule
\multicolumn{1}{c}{Condition} & Model                & QED                  & logP & SA Score \\ \midrule
-        & ZINC250k & $0.74 \pm 0.13$ & $2.45 \pm 1.44$ & $3.06 \pm 0.84$ \\ \cmidrule(l){2-5} 
\multirow{2}{*}{QED=0.7, logP=3.0, SAS=3}             & WGAN+Flow & $0.52 \pm 0.17$ & $5.43 \pm 1.89$ & $3.98 \pm 1.04$ \\
                              & Decaf & $0.68 \pm 0.10$ & $3.34 \pm 1.31$ & $3.07 \pm 0.29$  \\\midrule
\multirow{2}{*}{QED=0.5} & WGAN+Flow & $0.57 \pm 0.15$ & $4.66 \pm 1.67$ & $4.46 \pm 1.27$  \\ 
 & Decaf & $0.49 \pm 0.14$ & $3.63 \pm 1.65$ & $3.60 \pm 1.13$ \\ \cmidrule(l){2-5}
\multirow{2}{*}{QED=0.7} & WGAN+Flow & $0.53 \pm 0.18$ & $4.58 \pm 1.88$ & $3.95 \pm 1.01$  \\ 
 & Decaf & $0.67 \pm 0.13$&$3.22 \pm 1.55$&$4.14 \pm 1.18$\\ \cmidrule(l){2-5}
\multirow{2}{*}{QED=0.9} & WGAN+Flow & $0.60 \pm 0.13$ & $4.55 \pm 1.77$ & $5.06 \pm 1.35$ \\ 
 & Decaf &$0.77 \pm 0.08$ & $2.89 \pm 1.24$ & $5.27 \pm 0.93$\\\midrule
\multirow{2}{*}{logP=1.5} & WGAN+Flow & $0.64 \pm 0.15$ & $1.61 \pm 0.27$ & $4.43 \pm 1.28$  \\ 
 & Decaf &$0.60 \pm 0.16$&$1.91 \pm 1.68$&$4.45 \pm 1.12$\\   \cmidrule(l){2-5}
\multirow{2}{*}{logP=3.0} & WGAN+Flow & $0.65 \pm 0.12$ & $3.01 \pm 0.31$ & $4.06 \pm 1.48$  \\ 
 & Decaf & $0.54 \pm 0.18$&$3.17 \pm 1.91$&$4.57 \pm 1.28$\\ \cmidrule(l){2-5}
\multirow{2}{*}{logP=4.5} & WGAN+Flow & $0.58 \pm 0.14$ & $4.55 \pm 0.26$ & $4.28 \pm 1.39$ \\ 
 & Decaf & $0.60 \pm 0.12$ & $4.52 \pm 1.31$ & $5.44 \pm 1.17$\\  \midrule
\multirow{2}{*}{SAS=3} & WGAN+Flow & $0.44 \pm 0.16$ & $5.59 \pm 1.93$ & $3.11 \pm 0.26$ \\ 
 & Decaf & $0.59 \pm 0.14$&$3.70 \pm 1.69$&$3.73 \pm 1.36$ \\ \cmidrule(l){2-5}
\multirow{2}{*}{SAS=4} & WGAN+Flow & $0.57 \pm 0.16$ & $4.57 \pm 1.62$ & $3.95 \pm 0.29$ \\ 
 & Decaf &$0.58 \pm 0.17$ & $3.49 \pm 1.74$ & $4.36 \pm 1.14$\\  \cmidrule(l){2-5}
\multirow{2}{*}{SAS=5} & WGAN+Flow & $0.54 \pm 0.18$ & $4.87 \pm 1.67$ & $4.98 \pm 0.27$ \\ 
 & Decaf &$0.63 \pm 0.14$ & $2.34 \pm 1.75$ & $5.71 \pm 1.02$\\  \bottomrule
\end{tabular}
\caption{Conditional generation with WGAN and Decaf model.}
\end{table}

\begin{figure}
    \centering
    \includegraphics[width=\linewidth]{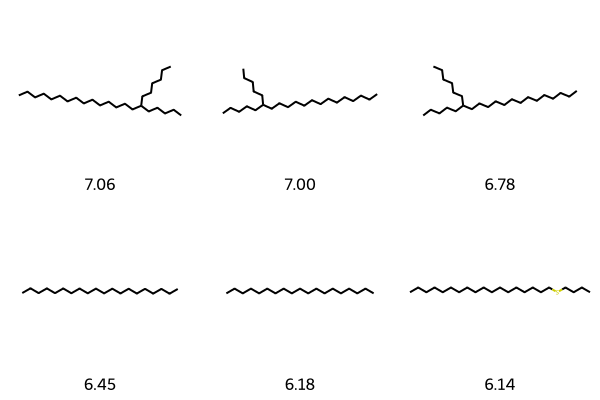}
    \caption{Molecules generated by unconstrained penalized logP generation from decaf.}
    \label{fig:unconst_plogp_mols}
\end{figure}

\begin{figure}
    \centering
    \includegraphics[width=\linewidth]{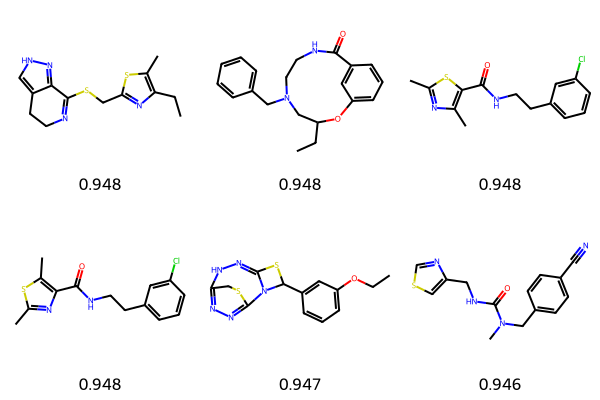}
    \caption{Molecules generated by unconstrained QED generation from decaf.}
    \label{fig:unconst_qed_mols}
\end{figure}

%% file: example_paper.bib
@article{mercatali2022symmetry,
  title={Symmetry-induced Disentanglement on Graphs},
  author={Mercatali, Giangiacomo and Freitas, Andr{\'e} and Garg, Vikas},
  journal={Advances in Neural Information Processing Systems},
  volume={35},
  pages={31497--31511},
  year={2022}
}

@inproceedings{zhu2021commutative,
  title={Commutative lie group vae for disentanglement learning},
  author={Zhu, Xinqi and Xu, Chang and Tao, Dacheng},
  booktitle={International Conference on Machine Learning},
  pages={12924--12934},
  year={2021},
  organization={PMLR}
}

@misc{styleganmnist,
author = {ritvik1909},
title = {Kaggle},
howpublished = {\url{https://www.kaggle.com/code/ritvik1909/style-gan/notebook}},
note = {2024},
year = {2022}
}

@article{ngo2022transitive,
  title={The Transitive Information Theory and its Application to Deep Generative Models},
  author={Ngo, Trung and Laabid, Najwa and Hautam{\"a}ki, Ville and Hein{\"a}niemi, Merja},
  journal={arXiv preprint arXiv:2203.05074},
  year={2022}
}

@article{kingma2013auto,
  title={Auto-encoding variational bayes},
  author={Kingma, Diederik P and Welling, Max},
  journal={arXiv preprint arXiv:1312.6114},
  year={2013}
}

@article{goodfellow2020generative,
  title={Generative adversarial networks},
  author={Goodfellow, Ian and Pouget-Abadie, Jean and Mirza, Mehdi and Xu, Bing and Warde-Farley, David and Ozair, Sherjil and Courville, Aaron and Bengio, Yoshua},
  journal={Communications of the ACM},
  volume={63},
  number={11},
  pages={139--144},
  year={2020},
  publisher={ACM New York, NY, USA}
}

@article{dinh2014nice,
  title={Nice: Non-linear independent components estimation},
  author={Dinh, Laurent and Krueger, David and Bengio, Yoshua},
  journal={arXiv preprint arXiv:1410.8516},
  year={2014}
}

@article{dinh2016density,
  title={Density estimation using real nvp},
  author={Dinh, Laurent and Sohl-Dickstein, Jascha and Bengio, Samy},
  journal={arXiv preprint arXiv:1605.08803},
  year={2016}
}

@article{mirza2014conditional,
  title={Conditional generative adversarial nets},
  author={Mirza, Mehdi and Osindero, Simon},
  journal={arXiv preprint arXiv:1411.1784},
  year={2014}
}

@inproceedings{odena2017conditional,
  title={Conditional image synthesis with auxiliary classifier gans},
  author={Odena, Augustus and Olah, Christopher and Shlens, Jonathon},
  booktitle={International conference on machine learning},
  pages={2642--2651},
  year={2017},
  organization={PMLR}
}

@inproceedings{zang2020moflow,
  title={MoFlow: an invertible flow model for generating molecular graphs},
  author={Zang, Chengxi and Wang, Fei},
  booktitle={Proceedings of the 26th ACM SIGKDD International Conference on Knowledge Discovery \& Data Mining},
  pages={617--626},
  year={2020}
}

@article{you2018graph,
  title={Graph convolutional policy network for goal-directed molecular graph generation},
  author={You, Jiaxuan and Liu, Bowen and Ying, Rex and Pande, Vijay and Leskovec, Jure},
  journal={arXiv preprint arXiv:1806.02473},
  year={2018}
}

@article{irwin2012zinc,
  title={ZINC: a free tool to discover chemistry for biology},
  author={Irwin, John J and Sterling, Teague and Mysinger, Michael M and Bolstad, Erin S and Coleman, Ryan G},
  journal={Journal of chemical information and modeling},
  volume={52},
  number={7},
  pages={1757--1768},
  year={2012},
  publisher={ACS Publications}
}

@inproceedings{liu2018constrained,
  title={Constrained graph variational autoencoders for molecule design},
  author={Liu, Qi and Allamanis, Miltiadis and Brockschmidt, Marc and Gaunt, Alexander},
  booktitle={Advances in neural information processing systems},
  pages={7795--7804},
  year={2018}
}

@inproceedings{bao2017cvae,
  title={CVAE-GAN: fine-grained image generation through asymmetric training},
  author={Bao, Jianmin and Chen, Dong and Wen, Fang and Li, Houqiang and Hua, Gang},
  booktitle={Proceedings of the IEEE international conference on computer vision},
  pages={2745--2754},
  year={2017}
}

@inproceedings{xian2019f,
  title={f-vaegan-d2: A feature generating framework for any-shot learning},
  author={Xian, Yongqin and Sharma, Saurabh and Schiele, Bernt and Akata, Zeynep},
  booktitle={Proceedings of the IEEE/CVF conference on computer vision and pattern recognition},
  pages={10275--10284},
  year={2019}
}

@article{lampert2013attribute,
  title={Attribute-based classification for zero-shot visual object categorization},
  author={Lampert, Christoph H and Nickisch, Hannes and Harmeling, Stefan},
  journal={IEEE transactions on pattern analysis and machine intelligence},
  volume={36},
  number={3},
  pages={453--465},
  year={2013},
  publisher={IEEE}
}

@inproceedings{liu2019conditional,
  title={Conditional adversarial generative flow for controllable image synthesis},
  author={Liu, Rui and Liu, Yu and Gong, Xinyu and Wang, Xiaogang and Li, Hongsheng},
  booktitle={Proceedings of the IEEE/CVF Conference on Computer Vision and Pattern Recognition},
  pages={7992--8001},
  year={2019}
}

@article{lippe2020categorical,
  title={Categorical normalizing flows via continuous transformations},
  author={Lippe, Phillip and Gavves, Efstratios},
  journal={arXiv preprint arXiv:2006.09790},
  year={2020}
}

@article{bengio2013representation,
  title={Representation learning: A review and new perspectives},
  author={Bengio, Yoshua and Courville, Aaron and Vincent, Pascal},
  journal={IEEE transactions on pattern analysis and machine intelligence},
  volume={35},
  number={8},
  pages={1798--1828},
  year={2013},
  publisher={IEEE}
}

@article{alemi2016deep,
  title={Deep variational information bottleneck},
  author={Alemi, Alexander A and Fischer, Ian and Dillon, Joshua V and Murphy, Kevin},
  journal={arXiv preprint arXiv:1612.00410},
  year={2016}
}

@article{burgess2018understanding,
	author = {Burgess, Christopher P and Higgins, Irina and Pal, Arka and Matthey, Loic and Watters, Nick and Desjardins, Guillaume and Lerchner, Alexander},
	journal = {NIPS Workshop on Learning Disentangled Representations},
	title = {Understanding disentangling in $\beta$-{VAE}},
	year = {2017}}

@inproceedings{chen2016infogan,
	author = {Chen, Xi and Duan, Yan and Houthooft, Rein and Schulman, John and Sutskever, Ilya and Abbeel, Pieter},
	booktitle = {{A}dvances in {N}eural {I}nformation {P}rocessing {S}ystems},
	pages = {2172--2180},
	title = {{I}nfogan: {I}nterpretable representation learning by information maximizing generative adversarial nets},
	year = {2016}}

@inproceedings{esmaeili2019structured,
	author = {Esmaeili, Babak and Wu, Hao and Jain, Sarthak and Bozkurt, Alican and Siddharth, Narayanaswamy and Paige, Brooks and Brooks, Dana H and Dy, Jennifer and Meent, Jan-Willem},
	booktitle = {The 22nd {I}nternational {C}onference on {A}rtificial {I}ntelligence and {S}tatistics},
	organization = {PMLR},
	pages = {2525--2534},
	title = {Structured disentangled representations},
	year = {2019}}

@inproceedings{kumar2018variational,
	author = {Abhishek Kumar and Prasanna Sattigeri and Avinash Balakrishnan},
	booktitle = {{I}nternational {C}onference on {L}earning {R}epresentations},
	title = {Variational {I}nference of {D}isentangled {L}atent {C}oncepts from {U}nlabeled {O}bservations},
	url = {https://openreview.net/forum?id=H1kG7GZAW},
	year = {2018}}

@inproceedings{mathieu2019disentangling,
	author = {Mathieu, Emile and Rainforth, Tom and Siddharth, N and Teh, Yee Whye},
	booktitle = {{I}nternational {C}onference on {M}achine {L}earning},
	pages = {4402--4412},
	title = {Disentangling disentanglement in variational autoencoders},
	year = {2019}}

@inproceedings{rolinek2019variational,
	author = {Rolinek, Michal and Zietlow, Dominik and Martius, Georg},
	booktitle = {Proceedings {IEEE} {C}onf. on {C}omputer {V}ision and {P}attern {R}ecognition},
	title = {Variational autoencoders recover {PCA} directions (by accident)},
	year = {2019}}

@misc{rubenstein2018learning,
	author = {Paul K. Rubenstein and Bernhard Schoelkopf and Ilya Tolstikhin},
	journal = {ICLR Workshops},
	title = {Learning {D}isentangled {R}epresentations with {W}asserstein {A}uto-{E}ncoders},
	year = {2018}}

@misc{
mollaysa2020conditional,
title={Conditional generation of molecules from disentangled representations},
author={Amina Mollaysa and Brooks Paige and Alexandros  Kalousis},
year={2020},
url={https://openreview.net/forum?id=BkxthxHYvr}
}

@inproceedings{jeon2021ib,
  title={Ib-gan: Disentangled representation learning with information bottleneck generative adversarial networks},
  author={Jeon, Insu and Lee, Wonkwang and Pyeon, Myeongjang and Kim, Gunhee},
  booktitle={Proceedings of the AAAI Conference on Artificial Intelligence},
  volume={35},
  pages={7926--7934},
  year={2021}
}

@inproceedings{nie2020semi,
  title={Semi-supervised stylegan for disentanglement learning},
  author={Nie, Weili and Karras, Tero and Garg, Animesh and Debnath, Shoubhik and Patney, Anjul and Patel, Ankit and Anandkumar, Animashree},
  booktitle={International Conference on Machine Learning},
  pages={7360--7369},
  year={2020},
  organization={PMLR}
}

@inproceedings{lin2020infogan,
  title={Infogan-cr and modelcentrality: Self-supervised model training and selection for disentangling gans},
  author={Lin, Zinan and Thekumparampil, Kiran and Fanti, Giulia and Oh, Sewoong},
  booktitle={international conference on machine learning},
  pages={6127--6139},
  year={2020},
  organization={PMLR}
}

@article{kang2018conditional,
  title={Conditional molecular design with deep generative models},
  author={Kang, Seokho and Cho, Kyunghyun},
  journal={Journal of chemical information and modeling},
  volume={59},
  number={1},
  pages={43--52},
  year={2018},
  publisher={ACS Publications}
}

@article{lecun1998mnist,
  title={The MNIST database of handwritten digits},
  author={LeCun, Yann},
  journal={http://yann. lecun. com/exdb/mnist/},
  year={1998}
}

@inproceedings{
higgins2017betavae,
title={beta-{VAE}: Learning Basic Visual Concepts with a Constrained Variational Framework},
author={Irina Higgins and Loic Matthey and Arka Pal and Christopher Burgess and Xavier Glorot and Matthew Botvinick and Shakir Mohamed and Alexander Lerchner},
booktitle={International Conference on Learning Representations},
year={2017},
url={https://openreview.net/forum?id=Sy2fzU9gl}
}

@misc{heusel2018gans,
      title={GANs Trained by a Two Time-Scale Update Rule Converge to a Local Nash Equilibrium}, 
      author={Martin Heusel and Hubert Ramsauer and Thomas Unterthiner and Bernhard Nessler and Sepp Hochreiter},
      year={2018},
      eprint={1706.08500},
      archivePrefix={arXiv},
      primaryClass={cs.LG}
}

@inproceedings{kim2018disentangling,
  title={Disentangling by factorising},
  author={Kim, Hyunjik and Mnih, Andriy},
  booktitle={International Conference on Machine Learning},
  pages={2649--2658},
  year={2018},
  organization={PMLR}
}

@inproceedings{
eastwood2018a,
title={A framework for the quantitative evaluation of disentangled representations},
author={Cian Eastwood and Christopher K. I. Williams},
booktitle={International Conference on Learning Representations},
year={2018},
url={https://openreview.net/forum?id=By-7dz-AZ},
}

@article{kumar2017variational,
  title={Variational inference of disentangled latent concepts from unlabeled observations},
  author={Kumar, Abhishek and Sattigeri, Prasanna and Balakrishnan, Avinash},
  journal={arXiv preprint arXiv:1711.00848},
  year={2017}
}

@article{chen2018isolating,
  title={Isolating sources of disentanglement in variational autoencoders},
  author={Chen, Ricky TQ and Li, Xuechen and Grosse, Roger B and Duvenaud, David K},
  journal={Advances in neural information processing systems},
  volume={31},
  year={2018}
}

@inproceedings{jin2018junction,
  title={Junction tree variational autoencoder for molecular graph generation},
  author={Jin, Wengong and Barzilay, Regina and Jaakkola, Tommi},
  booktitle={International conference on machine learning},
  pages={2323--2332},
  year={2018},
  organization={PMLR}
}

@article{verma2022modular,
  title={Modular flows: Differential molecular generation},
  author={Verma, Yogesh and Kaski, Samuel and Heinonen, Markus and Garg, Vikas},
  journal={Advances in Neural Information Processing Systems},
  volume={35},
  pages={12409--12421},
  year={2022}
}

@article{shi2020graphaf,
  title={Graphaf: a flow-based autoregressive model for molecular graph generation},
  author={Shi, Chence and Xu, Minkai and Zhu, Zhaocheng and Zhang, Weinan and Zhang, Ming and Tang, Jian},
  journal={arXiv preprint arXiv:2001.09382},
  year={2020}
}

@inproceedings{luo2021graphdf,
  title={Graphdf: A discrete flow model for molecular graph generation},
  author={Luo, Youzhi and Yan, Keqiang and Ji, Shuiwang},
  booktitle={International Conference on Machine Learning},
  pages={7192--7203},
  year={2021},
  organization={PMLR}
}

@misc{ho2019flow,
      title={Flow++: Improving Flow-Based Generative Models with Variational Dequantization and Architecture Design}, 
      author={Jonathan Ho and Xi Chen and Aravind Srinivas and Yan Duan and Pieter Abbeel},
      year={2019},
      eprint={1902.00275},
      archivePrefix={arXiv},
      primaryClass={cs.LG}
}

@article{ramakrishnan2014quantum,
  title={Quantum chemistry structures and properties of 134 kilo molecules},
  author={Ramakrishnan, Raghunathan and Dral, Pavlo O and Rupp, Matthias and von Lilienfeld, O Anatole},
  journal={Scientific Data},
  volume={1},
  year={2014},
  publisher={Nature Publishing Group}
}

@article{10.3389/fphar.2020.565644,
  title={{M}olecular {S}ets ({MOSES}): {A} {B}enchmarking {P}latform for {M}olecular {G}eneration {M}odels},
  author={Polykovskiy, Daniil and Zhebrak, Alexander and Sanchez-Lengeling, Benjamin and Golovanov, Sergey and Tatanov, Oktai and Belyaev, Stanislav and Kurbanov, Rauf and Artamonov, Aleksey and Aladinskiy, Vladimir and Veselov, Mark and Kadurin, Artur and Johansson, Simon and  Chen, Hongming and Nikolenko, Sergey and Aspuru-Guzik, Alan and Zhavoronkov, Alex},
  journal={Frontiers in Pharmacology},
  year={2020}
}

@software{landrum_2015_10398,
	author = {Landrum, Gregory},
	doi = {10.5281/zenodo.10398},
	month = jun,
	publisher = {Zenodo},
	title = {{RDKit: Open-source cheminformatics. Release 2014.03.1}},
	url = {https://doi.org/10.5281/zenodo.10398},
	year = 2015,}

@article{ertl2009estimation,
  title={Estimation of synthetic accessibility score of drug-like molecules based on molecular complexity and fragment contributions},
  author={Ertl, Peter and Schuffenhauer, Ansgar},
  journal={Journal of cheminformatics},
  volume={1},
  pages={1--11},
  year={2009},
  publisher={Springer}
}

@article{zhang2023survey,
  title={A survey of controllable text generation using transformer-based pre-trained language models},
  author={Zhang, Hanqing and Song, Haolin and Li, Shaoyu and Zhou, Ming and Song, Dawei},
  journal={ACM Computing Surveys},
  volume={56},
  number={3},
  pages={1--37},
  year={2023},
  publisher={ACM New York, NY}
}

@article{croitoru2023diffusion,
  title={Diffusion models in vision: A survey},
  author={Croitoru, Florinel-Alin and Hondru, Vlad and Ionescu, Radu Tudor and Shah, Mubarak},
  journal={IEEE Transactions on Pattern Analysis and Machine Intelligence},
  year={2023},
  publisher={IEEE}
}

@article{xue2019advances,
  title={Advances and challenges in deep generative models for de novo molecule generation},
  author={Xue, Dongyu and Gong, Yukang and Yang, Zhaoyi and Chuai, Guohui and Qu, Sheng and Shen, Aizong and Yu, Jing and Liu, Qi},
  journal={Wiley Interdisciplinary Reviews: Computational Molecular Science},
  volume={9},
  number={3},
  pages={e1395},
  year={2019},
  publisher={Wiley Online Library}
}

@article{pirhayatifard2023improving,
  title={Improving Denoising Diffusion Probabilistic Models via Exploiting Shared Representations},
  author={Pirhayatifard, Delaram and Toghani, Mohammad Taha and Balakrishnan, Guha and Uribe, C{\'e}sar A},
  journal={arXiv preprint arXiv:2311.16353},
  year={2023}
}

@article{verma2023abode,
  title={AbODE: Ab Initio Antibody Design using Conjoined ODEs},
  author={Verma, Yogesh and Heinonen, Markus and Garg, Vikas},
  journal={arXiv preprint arXiv:2306.01005},
  year={2023}
}

@article{song2020score,
  title={Score-based generative modeling through stochastic differential equations},
  author={Song, Yang and Sohl-Dickstein, Jascha and Kingma, Diederik P and Kumar, Abhishek and Ermon, Stefano and Poole, Ben},
  journal={arXiv preprint arXiv:2011.13456},
  year={2020}
}

@article{lee2021learning,
  title={Learning debiased representation via disentangled feature augmentation},
  author={Lee, Jungsoo and Kim, Eungyeup and Lee, Juyoung and Lee, Jihyeon and Choo, Jaegul},
  journal={Advances in Neural Information Processing Systems},
  volume={34},
  pages={25123--25133},
  year={2021}
}

@article{wu2020improving,
  title={Improving robustness and generality of nlp models using disentangled representations},
  author={Wu, Jiawei and Li, Xiaoya and Ao, Xiang and Meng, Yuxian and Wu, Fei and Li, Jiwei},
  journal={arXiv preprint arXiv:2009.09587},
  year={2020}
}

@article{van2019disentangled,
  title={Are disentangled representations helpful for abstract visual reasoning?},
  author={Van Steenkiste, Sjoerd and Locatello, Francesco and Schmidhuber, J{\"u}rgen and Bachem, Olivier},
  journal={Advances in neural information processing systems},
  volume={32},
  year={2019}
}

@misc{he2015deep,
      title={Deep Residual Learning for Image Recognition}, 
      author={Kaiming He and Xiangyu Zhang and Shaoqing Ren and Jian Sun},
      year={2015},
      eprint={1512.03385},
      archivePrefix={arXiv},
      primaryClass={cs.CV}
}

@misc{arjovsky2017wasserstein,
      title={Wasserstein GAN}, 
      author={Martin Arjovsky and Soumith Chintala and Léon Bottou},
      year={2017},
      eprint={1701.07875},
      archivePrefix={arXiv},
      primaryClass={stat.ML}
}

@inproceedings{karras2020analyzing,
  title={Analyzing and improving the image quality of stylegan},
  author={Karras, Tero and Laine, Samuli and Aittala, Miika and Hellsten, Janne and Lehtinen, Jaakko and Aila, Timo},
  booktitle={Proceedings of the IEEE/CVF conference on computer vision and pattern recognition},
  pages={8110--8119},
  year={2020}
}

@article{dhariwal2021diffusion,
  title={Diffusion models beat gans on image synthesis},
  author={Dhariwal, Prafulla and Nichol, Alexander},
  journal={Advances in neural information processing systems},
  volume={34},
  pages={8780--8794},
  year={2021}
}

@article{ho2022classifier,
  title={Classifier-free diffusion guidance},
  author={Ho, Jonathan and Salimans, Tim},
  journal={arXiv preprint arXiv:2207.12598},
  year={2022}
}

@article{sinha2021d2c,
  title={D2c: Diffusion-decoding models for few-shot conditional generation},
  author={Sinha, Abhishek and Song, Jiaming and Meng, Chenlin and Ermon, Stefano},
  journal={Advances in Neural Information Processing Systems},
  volume={34},
  pages={12533--12548},
  year={2021}
}

@INPROCEEDINGS{10447350,
  author={Ma, Changsheng and Guo, Taicheng and Yang, Qiang and Chen, Xiuying and Gao, Xin and Liang, Shangsong and Chawla, Nitesh and Zhang, Xiangliang},
  booktitle={ICASSP 2024 - 2024 IEEE International Conference on Acoustics, Speech and Signal Processing (ICASSP)}, 
  title={A Property-Guided Diffusion Model For Generating Molecular Graphs}, 
  year={2024},
  volume={},
  number={},
  pages={2365-2369},
  doi={10.1109/ICASSP48485.2024.10447350}}

@article{yang2023disdiff,
  title={Disdiff: Unsupervised disentanglement of diffusion probabilistic models},
  author={Yang, Tao and Wang, Yuwang and Lv, Yan and Zheng, Nanning},
  journal={arXiv preprint arXiv:2301.13721},
  year={2023}
}

@inproceedings{wu2023uncovering,
  title={Uncovering the disentanglement capability in text-to-image diffusion models},
  author={Wu, Qiucheng and Liu, Yujian and Zhao, Handong and Kale, Ajinkya and Bui, Trung and Yu, Tong and Lin, Zhe and Zhang, Yang and Chang, Shiyu},
  booktitle={Proceedings of the IEEE/CVF conference on computer vision and pattern recognition},
  pages={1900--1910},
  year={2023}
}
